\documentclass[pdflatex,iicol,sn-basic]{sn-jnl}

\usepackage{graphicx}%
\usepackage{multirow}%
\usepackage{amsmath,amssymb,amsfonts}%
\usepackage{amsthm}%
\usepackage[title]{appendix}%
\usepackage{xcolor}%
\usepackage{textcomp}%
\usepackage{manyfoot}%
\usepackage{booktabs}%
\usepackage{algorithm}%
\usepackage{algorithmicx}%
\usepackage{algpseudocode}%
\usepackage{listings}%
\usepackage{array}%
\usepackage{xspace}%
\usepackage{bookmark}%
\definecolor{orcidgreen}{HTML}{A6CE39}%
\newcommand{\orcidbadge}{\raisebox{0.2ex}{\scalebox{0.75}{\setlength{\fboxsep}{1.2pt}\colorbox{orcidgreen}{\scriptsize\rmfamily\bfseries\textcolor{white}{iD}}}}}%
\newcommand{\orcidlink}[1]{\href{https://orcid.org/#1}{\orcidbadge}}%

\newcolumntype{L}[1]{>{\raggedright\arraybackslash}p{#1}}
\newcommand{\method}{\textit{SelfThinker}\xspace}
\newcommand{\vt}{v}
\newcommand{\vmis}{v_{\mathrm{mis}}}

\definecolor{stgreen}{HTML}{009E73}
\definecolor{stred}{HTML}{C0392B}
\newcommand{\cmark}{\textcolor{stgreen}{\ensuremath{\boldsymbol{\checkmark}}}}
\newcommand{\xmark}{\textcolor{stred}{\ensuremath{\boldsymbol{\times}}}}


\begin{document}

\title[SelfThinker]{\method: Self-Evolving Reasoning via Reliability-Verified Intrinsic Visual Thoughts}

\author[1]{\fnm{Zhangquan} \sur{Chen}\,\orcidlink{0009-0006-2370-6120}}
\author*[1]{\fnm{Ruqi} \sur{Huang}\,\orcidlink{0000-0001-5942-3671}}\email{ruqihuang@sz.tsinghua.edu.cn}
\makeatletter
\g@addto@macro\corrauthemail{\unskip\par\smallskip\noindent\normalfont ORCID(s): \href{https://orcid.org/0009-0006-2370-6120}{\textcolor{blue}{0009-0006-2370-6120}} (Z.~Chen); \href{https://orcid.org/0000-0001-5942-3671}{\textcolor{blue}{0000-0001-5942-3671}} (R.~Huang)}
\makeatother

\affil[1]{\orgdiv{Tsinghua Shenzhen International Graduate School}, \orgname{Tsinghua University}, \orgaddress{\city{Shenzhen}, \postcode{518055}, \state{Guangdong}, \country{China}}}

\abstract{Unified multimodal models can both generate and understand images within a single set of weights, offering an opportunity to use image synthesis as training supervision for visual comprehension. However, existing visual chain-of-thought methods rely on human-annotated helper images during training and require slow image generation at test time. We present \method, \emph{the first self-evolving framework that turns a unified model's self-generated visual thoughts into stronger understanding without human image annotations.} Given a visual problem, \method samples candidate helper images from its generation pathway and uses its frozen understanding pathway to score each candidate via conditional pointwise mutual information. High-utility visual thoughts then refine the generator via flow matching and serve as privileged teacher context for on-policy self-distillation. At inference, \method reasons directly from the original image in text without generating intermediate images. Although trained only on small synthetic visual-search and navigation datasets, \method transfers broadly to natural-image and spatial benchmarks, raising the seven-benchmark average of ThinkMorph-7B from 52.7 to 64.0 (\textbf{+11.3 points}). Controlled image-perturbation probes further show that \method surpasses ThinkMorph-7B by up to \textbf{40.4 points} and its pretrained BAGEL-7B backbone by up to \textbf{23.2 points} across fine-grained and spatial perception tasks. Our code, data, and models are available at \url{https://github.com/zhangquanchen/SelfThinker}.}


\maketitle

\begin{figure*}[t]
\centering
\includegraphics[width=\textwidth]{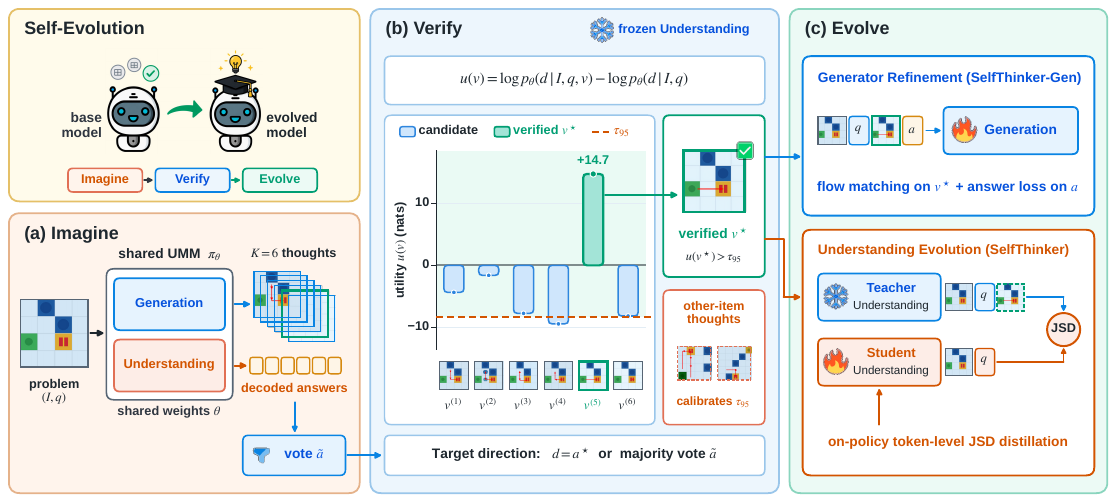}
\caption{Overview of \method. The upper-left \emph{Self-Evolution} panel summarizes the closed loop of imagining candidate visual thoughts, verifying reliable ones, and evolving into a stronger model. \emph{Imagine}: a shared UMM with Generation and Understanding pathways produces candidate visual thoughts and decoded answers. \emph{Verify}: the frozen Understanding pathway scores each candidate by its log-likelihood utility, calibrates the acceptance threshold against thoughts from other problems, and selects the top verified thought under the ground-truth answer or majority vote. \emph{Evolve}: the verified thought is distilled into the Student Understanding pathway via on-policy token-level JSD distillation from the EMA Teacher (\method), supported by flow-matching generator refinement (\method-Gen). No human image annotation is used}\label{fig:overview}
\end{figure*}

\section{Introduction}\label{sec:intro}

Visual understanding and spatial reasoning are foundational capabilities for multimodal large language models \citep{tong2024eyes,tong2024cambrian,fu2024blink}. When tackling multi-step visual tasks (e.g., grid navigation, visual search), text-only chains of thought often fall short because natural language struggles to capture fine-grained spatial geometry. Humans naturally overcome this bottleneck by forming intermediate mental images to guide deduction \citep{shepard1971mental,kosslyn2001neural,pearson2015mental,chen20253dthink,chen20264dthinker}. Motivated by this cognitive process, recent unified multimodal models (UMMs) integrate image generation and visual understanding within a single set of weights \citep{chameleon2024,zhou2024transfusion,xie2025showo,wu2024janus,wang2024emu3,tong2024metamorph,deng2025bagel}, enabling interleaved reasoning models to draw intermediate helper images, or visual thoughts, before predicting a final answer \citep{li2025mvot,chern2025thinking,gu2025thinkmorph,chen2025visrl,chen2025sifthinker}.

In principle, because a unified model can already synthesize images, its generation pathway could serve as a self-contained source of visual supervision to strengthen its understanding pathway. In practice, however, existing visual-thought pipelines such as Mirage \citep{yang2025machine}, ThinkMorph \citep{gu2025thinkmorph}, and Visual-OPSD \citep{li2026visualopsd} remain tied to external visual supervision, relying on human-designed helper images rendered by external tools (such as shortest-path overlays for navigation or bounding boxes for visual search). Constructing such task-specific templates requires costly manual engineering and confines visual imagination to predefined heuristics, \emph{preventing the model from learning from its own generated imagery at scale}.

Overcoming this limitation and turning a model's image generation capability into effective training supervision for visual understanding requires addressing three core challenges: \textbf{(C1) Annotation-free learning}, replacing human-engineered visual templates with diverse candidate visual thoughts sampled directly from the model's own generation pathway, without ground-truth bounding boxes, masks, or path overlays; \textbf{(C2) Intrinsic verification}, determining which self-generated images genuinely aid comprehension, since generated candidates are often noisy and standard majority voting over text answers \citep{wang2023selfconsistency,huang2023selfimprove,zuo2025ttrl,li2026uopsd,gkountouras2026canon} fails when agreeing answers accompany uninformative or corrupted intermediate images \citep{li2026visualopsd}; and \textbf{(C3) Test-time efficiency}, internalizing the visual cues of helpful intermediate images back into the understanding pathway, so that deployment neither incurs slow image generation latency nor halts prematurely at the image token.

To address (C1)--(C3), we propose \method (Fig.~\ref{fig:overview}), a self-evolving framework structured into three stages: \emph{Imagine}, \emph{Verify}, and \emph{Evolve}. In the \emph{Imagine} stage, given a visual problem, the model samples multiple candidate visual thoughts from its generation pathway and decodes their corresponding answers, supplying diverse visual candidates without human image annotations (resolving C1). In the \emph{Verify} stage, the model's frozen understanding pathway acts as an intrinsic verifier, scoring each candidate image by its conditional pointwise mutual information with a target answer direction and calibrating the score against thoughts drawn for other problems to filter out conflicting imagery (resolving C2). In the \emph{Evolve} stage, the verified visual thoughts serve as privileged teacher context for on-policy self-distillation (\method), transferring their visual cues into the understanding pathway so the student model reasons directly from the original image in text at test time (resolving C3), while a supporting flow-matching update (\method-Gen) refines the generator to produce cleaner candidate imagery.

Beyond eliminating human image annotations and test-time drawing overhead, our experiments also reveal a striking property, i.e., \emph{self-generated visual supervision transfers broadly across tasks}. Although \method is trained only on small synthetic visual-search and grid-navigation pools, its improvements generalize across seven diverse multimodal benchmarks, spanning fine-grained chart reading and visual search, multi-step spatial planning, and natural-scene perception. Specifically, \method raises the seven-benchmark average of ThinkMorph-7B from 52.7 to 64.0 ($+11.3$ points) under ground-truth answer supervision, and to 63.4 ($+10.7$ points) under label-free majority voting. Controlled image-perturbation probes further confirm that this cross-task transfer stems from stronger visual perception rather than output formatting or language priors. That is, \method consistently outperforms teacher-context controls that lack verified imagery, surpassing ThinkMorph-7B by up to 40.4 points and its pretrained BAGEL-7B-MoT backbone by 12.0 to 23.2 points across fine-grained and spatial benchmarks.

% Our main contributions are summarized as follows:
% \begin{itemize}
% \item We propose \method, the first self-evolving framework that turns a unified multimodal model's image generation capability into training supervision for visual understanding without human image annotations.
% \item We introduce an intrinsic visual verification mechanism based on calibrated conditional pointwise mutual information that selects informative self-generated helper images, and internalize them into text-only inference via privileged on-policy self-distillation.
% \item \method demonstrates strong cross-task transfer from synthetic training pools to natural-image and spatial perception.
% \end{itemize}

Our contributions can be summarized as follows.
\begin{itemize}
\item We propose \method, the first self-evolving framework that turns a unified multimodal model's image generation capability into training supervision for visual understanding without human image annotations.
\item We introduce an intrinsic visual verification mechanism based on calibrated conditional pointwise mutual information, enabling the understanding pathway to select informative self-generated helper images.
\item We develop a privileged on-policy self-distillation scheme, supported by flow-matching generator refinement, that internalizes verified visual thoughts into the understanding pathway for fast text-only inference without test-time image generation.
\item Extensive experiments across multiple benchmarks and controlled image-perturbation probes demonstrate that \method consistently outperforms strong baselines and its pretrained backbone. Furthermore, our results indicate that \method generalizes effectively from synthetic training pools to natural-image and spatial perception, highlighting its broad applicability.
\end{itemize}

% Our main contributions are summarized as follows:
% \begin{itemize}
% \item We propose \method, the first self-evolving framework that turns a unified multimodal model's image generation capability into training supervision for visual understanding without human image annotations.
% \item We introduce an intrinsic visual verification mechanism based on calibrated conditional pointwise mutual information to select informative self-generated helper images under both ground-truth answers and label-free majority votes, and internalize them into text-only inference via privileged on-policy self-distillation.
% \item We demonstrate across seven benchmarks and controlled image-perturbation probes that \method transfers broadly from synthetic training pools to natural-image and 2D/3D spatial perception, improving the average score of ThinkMorph-7B by $+11.3$ points and surpassing the pretrained BAGEL-7B backbone by up to 23.2 points.
% \end{itemize}

\section{Related Work}\label{sec:related}

\subsection{Unified Multimodal Models}
Unified multimodal models (UMMs) integrate visual understanding and image generation within a single architecture. Early-fusion autoregressive models such as Chameleon \citep{chameleon2024} and Emu3 \citep{wang2024emu3} represent both modalities as discrete token streams. In contrast, Transfusion \citep{zhou2024transfusion} and Show-o \citep{xie2025showo} combine next-token prediction with continuous diffusion or discrete denoising. Janus and Janus-Pro decouple the visual encoders used for reading and generation \citep{wu2024janus,chen2025januspro}, whereas MetaMorph aligns an understanding backbone with visual generation via instruction tuning \citep{tong2024metamorph}. BAGEL routes understanding and generation tokens through specialized transformer experts pretrained on interleaved multimodal data \citep{deng2025bagel}. At the evaluation level, standard multimodal benchmarks often suffer from language shortcuts where questions can be answered without inspecting the image \citep{chen2024mmstar}. This limitation has motivated vision-centric benchmarks such as MMVP, CV-Bench, and BlindTest \citep{tong2024eyes,tong2024cambrian,rahmanzadehgervi2024blind}, alongside image corruption robustness protocols \citep{hendrycks2019robustness}. \method builds on a unified architecture to convert its generative capability into stronger visual perception across fine-grained and spatial benchmarks.

\subsection{Reasoning with Helper Images}
\begin{table*}[t]
\caption{Comparison of \method with related visual-thought and self-distillation methods across five key properties: reasoning with intermediate visual thoughts, learning from the model's own self-generated visual thoughts rather than external tools or curated datasets, operating without human image annotations (bounding boxes, masks, or path overlays), explicitly verifying candidate visual thoughts, and supporting text-only inference at test time}\label{tab:positioning}
\footnotesize
\setlength{\tabcolsep}{3pt}
\begin{tabular*}{\textwidth}{@{\extracolsep\fill}lccccc@{}}
\toprule
Method & \begin{tabular}[c]{@{}c@{}}Visual\\thoughts\end{tabular} & \begin{tabular}[c]{@{}c@{}}Self-generated\\imagery\end{tabular} & \begin{tabular}[c]{@{}c@{}}No image\\annotations\end{tabular} & \begin{tabular}[c]{@{}c@{}}Thought\\verification\end{tabular} & \begin{tabular}[c]{@{}c@{}}Text-only\\inference\end{tabular} \\
\midrule
Visual Sketchpad \citep{hu2024sketchpad} & \cmark & \xmark & \xmark & \xmark & \xmark \\
MVoT \citep{li2025mvot}, ThinkMorph \citep{gu2025thinkmorph} & \cmark & \cmark & \xmark & \xmark & \xmark \\
Visual-OPSD \citep{li2026visualopsd}, Vision-OPD \citep{yuan2026visionopd} & \cmark & \xmark & \xmark & \xmark & \cmark \\
u-OPSD \citep{li2026uopsd}, Canon \citep{gkountouras2026canon} & \xmark & \xmark & \cmark & \xmark & \cmark \\
\midrule
\textbf{\method (ours)} & \cmark & \cmark & \cmark & \cmark & \cmark \\
\botrule
\end{tabular*}
\end{table*}

To move beyond text-only chains of thought, recent multimodal reasoning methods construct intermediate helper images that supply explicit visual cues during deduction. Tool-augmented systems such as Visual Sketchpad and Whiteboard-of-Thought invoke external code and drawing APIs to render auxiliary lines, boxes, and diagrams \citep{hu2024sketchpad,menon2024whiteboard}, while region-focusing methods crop or zoom into informative sub-regions \citep{shao2024visualcot,zheng2025deepeyes,fan2025grit}. Mirage synthesizes task-specific helper images, such as path-annotated maps, and compresses them into latent visual tokens during training \citep{yang2025machine}. Within native UMMs, MVoT generates step-by-step spatial visualizations \citep{li2025mvot}, and \citet{chern2025thinking} sample and critique intermediate images at test time. ThinkMorph fine-tunes BAGEL on interleaved traces across human-annotated helper-image templates, including red bounding boxes for visual search, shortest-path overlays for navigation, and region highlights for chart reading \citep{gu2025thinkmorph}. However, these approaches rely on external rendering tools or human annotation pipelines to prescribe task-specific helper images. In contrast, \method uses the model's own generation pathway as a source of visual training supervision without human image annotations, selects informative visual thoughts via the understanding pathway, and distills them into single-image text reasoning.

\subsection{Self-Evolving Training}
Self-evolving training frameworks iteratively improve model capability without human rationales by filtering self-generated trajectories \citep{zelikman2022star,gulcehre2023rest,dong2023raft,yuan2023scaling} or optimizing against outcome rewards and self-play \citep{chen2024spin,deepseek2025r1}. Related work also evaluates whether intermediate steps genuinely contribute to predictions via pointwise mutual information (PMI) and counterfactual interventions \citep{church1990pmi,koh2017influence,lanham2023faithfulness,turpin2023unfaithful}. A complementary paradigm is learning using privileged information \citep{vapnik2009lupi,hinton2015distilling,lopezpaz2016unifying}, which supplies a teacher with training signals that are unavailable to the student at test time. Combining privileged context with on-policy context distillation \citep{askell2021general,snell2022context,agarwal2024gkd} allows a student policy to internalize rich teacher conditioning along its own sampled trajectories. Recent methods instantiate this teacher context with annotated helper images in Visual-OPSD \citep{li2026visualopsd}, image crops in Vision-OPD \citep{yuan2026visionopd}, annealed contexts in ATESD \citep{han2026atesd}, and hindsight skills in SEED \citep{wu2026seed}.

When ground-truth labels are unavailable during self-training, majority voting across rollouts provides a label-free supervision signal \citep{wang2023selfconsistency,huang2023selfimprove,zuo2025ttrl}. Recent self-distillation methods such as u-OPSD \citep{li2026uopsd} and Canon \citep{gkountouras2026canon} condition the teacher on a consensus text solution, while UniGame pits generation against understanding in an adversarial loop \citep{su2025unigame}. As summarized in Table~\ref{tab:positioning}, prior training pipelines either require human-annotated helper images and external crops (ThinkMorph, Visual-OPSD, Vision-OPD) or distill text-only consensus solutions without visual thoughts (u-OPSD, Canon). \method bridges this gap through a unified self-evolving training loop. Rather than distilling dataset-provided visual annotations, we use conditional PMI against mismatched thoughts to verify helper images synthesized by the model itself, and distill the verified imagery into fast text-only inference.

\section{Methodology}\label{sec:method}

\subsection{Problem Formulation}
A unified multimodal model (UMM) $\pi_\theta = (\pi_\theta^U, \pi_\theta^G)$ shares parameters across an autoregressive understanding pathway $\pi_\theta^U$, which encodes images and generates text tokens, and a continuous generation pathway $\pi_\theta^G$, which synthesizes images in latent space \citep{deng2025bagel,gu2025thinkmorph}. Given a visual reasoning problem $x=(I,q)$ consisting of an input image $I$ and a query $q$, the model operates in the \emph{Imagine} stage by sampling $K$ interleaved trajectories $\{\tau^{(k)}\}_{k=1}^{K}$:
\begin{equation}
\tau^{(k)} = \big(r_{\mathrm{pre}}^{(k)},\, \vt^{(k)},\, r_{\mathrm{post}}^{(k)},\, a^{(k)}\big).\label{eq:traj}
\end{equation}
Here $r_{\mathrm{pre}}^{(k)} \sim \pi_\theta^U(\cdot \mid I,q)$ is the initial text thought before the image token \texttt{<image\_start>}, $\vt^{(k)} \sim \pi_\theta^G(\cdot \mid I,q,r_{\mathrm{pre}}^{(k)})$ is a self-generated helper image (an \emph{intrinsic visual thought}), and $(r_{\mathrm{post}}^{(k)}, a^{(k)}) \sim \pi_\theta^U(\cdot \mid I,q,r_{\mathrm{pre}}^{(k)},\vt^{(k)})$ is the text continuation ending in a decoded answer $a^{(k)}$.

Directly training on self-generated trajectories faces two obstacles. First, a correct text answer $a^{(k)} = a^\star$ can arise despite an uninformative or corrupted helper image $\vt^{(k)}$ \citep{li2026visualopsd}. Second, running image generation $\pi_\theta^G$ at test time incurs substantial latency, whereas disabling it causes premature termination at \texttt{<image\_start>}. As illustrated in Fig.~\ref{fig:overview} and Algorithm~\ref{alg:audit}, \method addresses both issues through two subsequent stages. In the \emph{Verify} stage (Sect.~\ref{sec:utility}), the understanding pathway scores the task utility of each candidate helper image $\vt^{(k)}$ without human image annotations. In the \emph{Evolve} stage, the model internalizes the verified visual context into the understanding pathway via on-policy self-distillation (Sect.~\ref{sec:und-evolve}), supported by flow-matching refinement of the generation pathway (Sect.~\ref{sec:gen-evolve}).

\subsection{Intrinsic Visual Verification}\label{sec:utility}\label{sec:screening}\label{sec:directions}
\paragraph{Directed utility score.}
To evaluate whether a candidate helper image $\vt$ provides useful visual evidence for solving $(I,q)$, we use the frozen understanding pathway $\pi_\theta^U$ as an intrinsic verifier. Let $d = (d_1, \dots, d_L)$ be a token sequence encoding a target answer direction. We define the directed utility of $\vt$ as the log-likelihood gain of generating $d$ when $\vt$ is revealed relative to the plain problem context:
\begin{equation}
u(\vt) = \log p_\theta\big(d \mid C(I,q,\vt)\big) - \log p_\theta\big(d \mid C(I,q)\big),\label{eq:utility}
\end{equation}
where $\log p_\theta(d \mid C) = \sum_{\ell=1}^{L} \log \pi_\theta^U(d_\ell \mid C, d_{<\ell})$. Here $C(I,q) = [I,q]$ is the single-image problem context, and $C(I,q,\vt) = [I,q,\rho,\vt,\tau]$ appends a neutral context header $\rho$, the helper image $\vt$, and a fixed transition instruction $\tau$ (prompt templates in Appendix~\ref{app:impl}). Measured in nats, $u(\vt)$ equals the conditional pointwise mutual information $\mathrm{PMI}_\theta(\vt; d \mid I,q)$ \citep{church1990pmi} and requires only two forward passes without gradients.

Equation~\eqref{eq:utility} has three desirable properties for verification without human image annotations. First, the score is signed, rewarding helper images that increase the probability of $d$ ($u(\vt) > 0$) and penalizing conflicting imagery ($u(\vt) < 0$). Second, omitting the rollout text $(r_{\mathrm{pre}}^{(k)}, r_{\mathrm{post}}^{(k)})$ from $C(I,q,\vt)$ isolates the visual contribution of $\vt$; because all candidates for item $i$ share the same header $\rho$ and instruction $\tau$, any prompt offset cancels in within-item comparisons $u_i(\vt_i^{(k)}) - u_i(\vmis^{(k)})$. Third, the score requires no ground-truth helper images or handcrafted pixel rules.

\begin{algorithm}[t]
\caption{\method: Self-Evolving Training}\label{alg:audit}
\footnotesize
\begin{algorithmic}[1]
\Require Model $\pi_\theta=(\pi_\theta^U,\pi_\theta^G)$, problem pool $\mathcal{D}=\{(I_i,q_i)\}_{i=1}^{N}$, rollouts $K$, quantile $\alpha=0.95$
\Ensure Updated generator \method-Gen ($\pi_\theta^G$) and distilled student \method ($\pi_\theta^U$)
\State $\mathcal{D}_{\mathrm{ver}} \gets \emptyset$
\For{$i = 1, \dots, N$} \Comment{\textbf{Stages 1--2: Imagine \& Verify}}
  \State Sample traces $\{(r_{\mathrm{pre},i}^{(k)}, \vt_i^{(k)}, r_{\mathrm{post},i}^{(k)}, a_i^{(k)})\}_{k=1}^{K} \sim \pi_\theta(\cdot \mid I_i, q_i)$
  \State Set $d_i \gets a_i^\star$ (answer mode) or $\tilde a_i = \mathrm{mode}(\{a_i^{(k)}\}_{k=1}^{K})$ (vote mode)
  \For{$k = 1, \dots, K$}
    \State $u_{ik} \gets u_i(\vt_i^{(k)})$, \quad $s_{ik} \gets u_i(\vmis^{(k)})$ \Comment{Eq.~\eqref{eq:utility}}
  \EndFor
  \State $k^\star \gets \arg\max_{k} u_{ik}$, \quad $Q_i \gets s_{i,(\lfloor \alpha(K-1)\rfloor + 1)}$
  \If{$u_{ik^\star} > Q_i$}
    \State $\mathcal{D}_{\mathrm{ver}} \gets \mathcal{D}_{\mathrm{ver}} \cup \{(I_i, q_i, \vt_i^{(k^\star)}, d_i)\}$ \Comment{Eq.~\eqref{eq:accept}}
  \EndIf
\EndFor
\For{minibatch $\mathcal{B} \subset \mathcal{D}_{\mathrm{ver}}$} \Comment{\textbf{Stage 3: Evolve}}
  \State \method-Gen: update $\pi_\theta^G$ on $[I,q,\vt^\star,d]$ via $\nabla_\theta \mathcal{L}_{\mathrm{Gen}}(\theta)$ \Comment{Eq.~\eqref{eq:genloss}}
  \State \method: sample $y \sim \pi_\theta^U(\cdot \mid c_S)$, update $\pi_\theta^U$ via $\nabla_\theta \mathcal{L}_{\mathrm{Und}}(\theta)$, update EMA $\bar\theta$ \Comment{Eq.~\eqref{eq:opsd}}
\EndFor
\end{algorithmic}
\end{algorithm}

\paragraph{Within-item calibration.}
Because $\log p_\theta(d \mid \cdot)$ scales with the token length $L$ of $d$, we calibrate utilities within each item $i$ against mismatched thoughts. Specifically, we pair each candidate $\vt_i^{(k)}$ with a mismatched thought $\vmis^{(k)} = \vt_{(i+\delta)\bmod N}^{(k)}$ generated by the same model for a different problem, and evaluate $s_{ik} = u_i(\vmis^{(k)})$ under item $i$'s context and direction. A mismatched thought matches a genuine candidate in style, resolution, and token count while depicting an unrelated scene, which controls for baseline conditioning bias. Sorting $\{s_{i1},\dots,s_{iK}\}$ as $s_{i,(1)} \le \dots \le s_{i,(K)}$ defines an item-adaptive order-statistic threshold $Q_i = s_{i,(\lfloor \alpha(K-1)\rfloor + 1)}$ ($\alpha=0.95$; Appendix~\ref{app:impl}). The top candidate is verified whenever it exceeds $Q_i$:
\begin{equation}
\vt_i^\star = \arg\max_{k} u_i\big(\vt_i^{(k)}\big), \quad u_i(\vt_i^\star) > Q_i.\label{eq:accept}
\end{equation}

\paragraph{Target directions.}
We instantiate $d_i$ in two settings, neither of which uses human helper-image annotations. The \emph{answer direction} sets $d_i = a_i^\star$ using the ground-truth text answer. The label-free \emph{vote direction} sets $d_i = \tilde a_i = \mathrm{mode}(\{a_i^{(k)}\}_{k=1}^{K})$ using the rollout majority vote, measuring how strongly each candidate image visually supports the consensus answer without requiring image or answer labels.

\subsection{Supporting Generative Refinement}\label{sec:gen-evolve}\label{sec:learning}
Given the verified dataset $\mathcal{D}_{\mathrm{ver}} = \{(I_i,q_i,\vt_i^\star,d_i)\}$, we can refine the generation pathway $\pi_\theta^G$ via rectified flow matching \citep{lipman2023flow,liu2023rectified,deng2025bagel} (\method-Gen) to improve the quality of candidate visual thoughts. Let $z_1 = E_{\mathrm{VAE}}(\vt^\star)$ denote the clean latent tokens of the verified helper image encoded by the variational autoencoder (VAE), $z_0 \sim \mathcal{N}(0,\mathbf{I})$ Gaussian noise, and $z_t = (1-t)z_0 + t z_1$ their linear interpolation at $t \sim \mathcal{U}(0,1)$. The generation expert $v_\theta^G$ regresses the target velocity $z_1 - z_0$ jointly with a cross-entropy loss $\mathcal{L}_{\mathrm{CE}}(\theta)$ on the answer tokens:
\begin{equation}
\begin{split}
\mathcal{L}_{\mathrm{Gen}}(\theta) = \,&\lambda_{\mathrm{FM}}\,\mathbb{E}_{t, z_0}\Big[\big\| v_\theta^G(z_t, t \mid I, q, r_{\mathrm{pre}})\\
&- (z_1 - z_0) \big\|_2^2\Big] + \lambda_{\mathrm{CE}}\,\mathcal{L}_{\mathrm{CE}}(\theta),
\end{split}\label{eq:genloss}
\end{equation}
with $\lambda_{\mathrm{FM}} = 1.0$ and $\lambda_{\mathrm{CE}} = 0.25$ (Appendix~\ref{app:impl}). Filtering training targets by Eq.~\eqref{eq:accept} shifts the generative prior toward problem-consistent trajectories and localized target boxes without human image annotations.

\subsection{Internalizing Visual Supervision into Understanding}\label{sec:und-evolve}\label{sec:diagnostics}
To convert verified visual thoughts into stronger visual comprehension without test-time drawing overhead, our primary update (\method) distills the verified helper images back into the understanding pathway via privileged on-policy self-distillation \citep{agarwal2024gkd,li2026visualopsd}. For each verified item $(I,q,\vt^\star) \in \mathcal{D}_{\mathrm{ver}}$, we construct asymmetric student and teacher contexts:
\begin{equation}
c_S = [I, q], \qquad c_T = [I, q, \rho, \vt^\star, \tau].\label{eq:contexts}
\end{equation}
The student $\pi_\theta^U$ observes only the test-time context $c_S$ and samples an on-policy text completion $y = (y_1, \dots, y_T) \sim \pi_\theta^U(\cdot \mid c_S)$ without image generation. An exponential moving average (EMA) teacher $\pi_{\bar\theta}^U$ ($\bar\theta \gets \mu\bar\theta + (1-\mu)\theta$) observes the privileged context $c_T$ containing the verified helper image $\vt^\star$. The teacher is initialized from $\theta$ and evaluated with stop-gradient, whereas the Stage~2 verifier remains fixed at the base ThinkMorph-7B weights. Letting $P_t^T = \pi_{\bar\theta}^U(\cdot \mid c_T, y_{<t})$, $P_t^S = \pi_\theta^U(\cdot \mid c_S, y_{<t})$, and $M_t = \beta P_t^T + (1-\beta) P_t^S$, we minimize the token-level generalized Jensen--Shannon divergence (JSD) along $y$ using the Kullback--Leibler ($\mathrm{KL}$) divergence:
\begin{equation}
\begin{split}
\mathcal{L}_{\mathrm{Und}}(\theta) = \,&\mathbb{E}_{y\sim\pi_\theta^U(\cdot\mid c_S)} \sum_{t=1}^{T} \Big[ \beta\,\mathrm{KL}(P_t^T \parallel M_t)\\
&+ (1-\beta)\,\mathrm{KL}(P_t^S \parallel M_t) \Big].
\end{split}\label{eq:opsd}
\end{equation}
Minimizing Eq.~\eqref{eq:opsd} teaches the student to bypass the \texttt{<image\_start>} bottleneck and extract the spatial cues highlighted by $\vt^\star$ directly from the original image $I$. \emph{At test time, \method reasons purely in text from $c_S=[I,q]$ without generating intermediate images.}

\section{Experiments}\label{sec:exp}

\subsection{Experimental Setup}\label{sec:setup}

\paragraph{Compared models.}
We build on ThinkMorph-7B \citep{gu2025thinkmorph}, an interleaved reasoning model fine-tuned from BAGEL-7B-MoT \citep{deng2025bagel} (denoted \emph{BAGEL-7B}). The model routes understanding and generation tokens through specialized Qwen2.5 transformer experts \citep{qwen2025qwen25} paired with a SigLIP vision encoder \citep{zhai2023siglip}. All intrinsic utility scores are evaluated by the frozen understanding pathway of ThinkMorph-7B. We evaluate our self-distilled understanding models across training pools and supervision settings: \method (Nav) (distilled from verified navigation thoughts), \method (Mix) (distilled from combined navigation and visual-search thoughts), \method (Vote) (distilled from visual-search thoughts verified via label-free majority voting), and our primary model \method (distilled from visual-search thoughts verified via ground-truth answers). In our ablation studies (Sect.~\ref{sec:res-abl}), we also evaluate two teacher-context controls under identical visual-search training schedules. Specifically, \emph{Noise-OPSD} replaces the teacher's verified thought with uniform RGB noise, and \emph{Prompt-OPSD} omits the second image so the teacher observes only the problem and transition instruction. For the generation pathway, we compare our supporting generative update (\method-Gen) against random-thought fine-tuning (\emph{Rand-RFT}).

\paragraph{Synthesized problem pools.}
\begin{table*}[t]
\caption{Main results on seven multimodal benchmarks under a unified judge (Qwen3-30B-A3B-Instruct-2507) and text-only decoding. \method variants distill verified visual thoughts from synthesized navigation (\emph{Nav}), combined (\emph{Mix}), or visual-search pools (under label-free rollout \emph{Vote} or ground-truth answer direction) into single-image text reasoning. Bold marks the best score in each column and $\Delta$ gives the gain over ThinkMorph-7B. CV-B.: CV-Bench}\label{tab:main}
\footnotesize
\setlength{\tabcolsep}{4.8pt}
\begin{tabular*}{\textwidth}{@{\extracolsep\fill}L{0.21\textwidth}cccccccc r@{}}
\toprule
Method & VSP & ChartQA & V* & MMVP & SAT & BLINK & CV-B. & Average & $\Delta$ \\
\midrule
ThinkMorph-7B \citep{gu2025thinkmorph} & 1.2 & 52.7 & 53.9 & 79.0 & 55.3 & 53.6 & 73.4 & 52.7 & -- \\
\midrule
\method (Nav) & 2.0 & 78.1 & 63.4 & 80.3 & 58.0 & \textbf{60.3} & 78.0 & 60.0 & $+7.3$ \\
\method (Mix) & 5.5 & 74.7 & \textbf{70.2} & 79.3 & 54.7 & 58.0 & \textbf{83.5} & 60.8 & $+8.1$ \\
\method (Vote) & 13.0 & \textbf{81.9} & 67.0 & 80.0 & \textbf{60.0} & 59.3 & 82.4 & 63.4 & $+10.7$ \\
\textbf{\method} & \textbf{15.8} & 80.1 & \textbf{70.2} & \textbf{82.3} & \textbf{60.0} & 58.2 & 81.7 & \textbf{64.0} & \textbf{$+11.3$} \\
\textbf{$\Delta$ vs.\ ThinkMorph-7B} & \textbf{$+14.6$} & \textbf{$+27.4$} & \textbf{$+16.3$} & \textbf{$+3.3$} & \textbf{$+4.7$} & \textbf{$+4.6$} & \textbf{$+8.3$} & \textbf{$+11.3$} & -- \\
\botrule
\end{tabular*}
\end{table*}

\begin{figure*}[t]
\centering
\includegraphics[width=\textwidth]{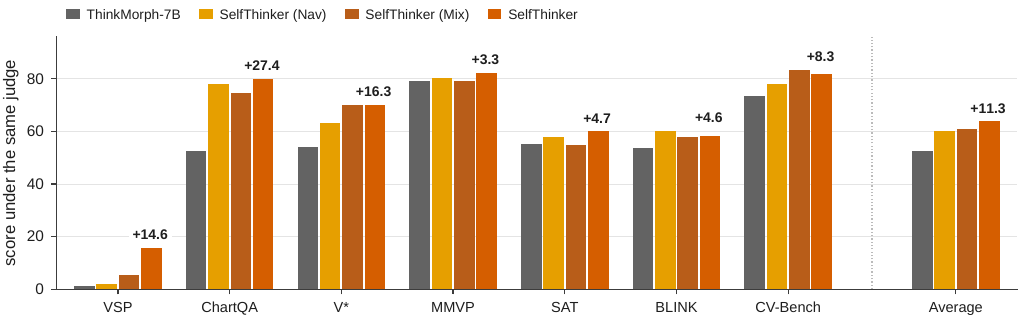}
\caption{Per-benchmark and average scores of ThinkMorph-7B, \method (Nav), \method (Mix), and \method under the unified judge and text-only decoding. Numbers above the bars indicate the gains of \method over ThinkMorph-7B}\label{fig:main}
\end{figure*}
To evaluate candidate visual thoughts on layouts unseen during ThinkMorph-7B training, we synthesize fresh \emph{Navigation} (Frozen-Lake grid maps \citep{towers2024gymnasium}) and \emph{Visual Search} (multi-object shape canvases) pools. Each pool is split into 1,500 training and 500 diagnostic items with disjoint layouts. For each problem, we sample $K=6$ interleaved traces with one visual thought per trace. Programmatic reference thoughts (shortest-path overlays and target boxes) are rendered only for post-hoc diagnostics and are never used in verification or training. Complete pool specifications, sampling parameters, and optimization schedules are provided in Appendix~\ref{app:impl}.

\paragraph{Evaluation protocols.}
Following the benchmark suite of \citet{li2026visualopsd}, we evaluate multimodal reasoning across seven benchmarks: VSP (600 maze items) \citep{wu2024vsp}, ChartQA (444 horizontal and 382 vertical bar items) \citep{masry2022chartqa}, V* (191 items) \citep{wu2024vstar}, MMVP (300 items) \citep{tong2024eyes}, SAT (150 circular split items) \citep{ray2024sat}, BLINK (1,901 items across 14 subtasks) \citep{fu2024blink}, and CV-Bench (1,438 2D and 1,200 3D items) \citep{tong2024cambrian}. All benchmarks are evaluated in VLMEvalKit \citep{duan2024vlmevalkit} using greedy text-only decoding (up to 1,024 thinking tokens) and a unified Qwen3-30B-A3B-Instruct-2507 judge \citep{yang2025qwen3,kwon2023vllm}. On the 500-item diagnostic splits, we evaluate intrinsic verification using four metrics: verification \emph{yield}, \emph{Top-1 AUC} (area under the receiver operating characteristic curve) \citep{hanley1982auc}, paired \emph{win rate}, and utility \emph{margin} (nats) (formal definitions in Appendix~\ref{app:defs}). Finally, to test whether accuracy gains stem from visual perception rather than language priors, we evaluate a multi-layer image-condition probe across image swaps, pixel scrambles, tile permutations, mosaics, and Gaussian blurs (Appendix~\ref{app:probe}).

\subsection{Main Benchmark Results}\label{sec:res-main}

\paragraph{Improving visual understanding without test-time image generation.}
Table~\ref{tab:main} and Fig.~\ref{fig:main} present the main benchmark comparison. By distilling verified visual thoughts into the understanding pathway, \emph{\method raises the seven-benchmark average from 52.7 to 64.0 ($+11.3$ points)}. It outperforms ThinkMorph-7B on every benchmark while operating purely in text mode at test time, avoiding the 59--75\,s per-trace diffusion overhead required by interleaved image generation. Notably, even though training uses only a small pool of synthetic navigation or visual-search problems, the distilled visual supervision transfers broadly across natural-image and spatial benchmarks. Specifically, \method (Nav) lifts the seven-benchmark average to 60.0 ($+7.3$ points, reaching 60.3 on BLINK), \method (Mix) reaches 60.8 ($+8.1$ points, reaching 83.5 on CV-Bench), label-free \method (Vote) reaches 63.4 ($+10.7$ points, reaching 81.9 on ChartQA), and \method reaches 64.0 ($+11.3$ points).

\begin{figure*}[t]
\centering
\includegraphics[width=\textwidth]{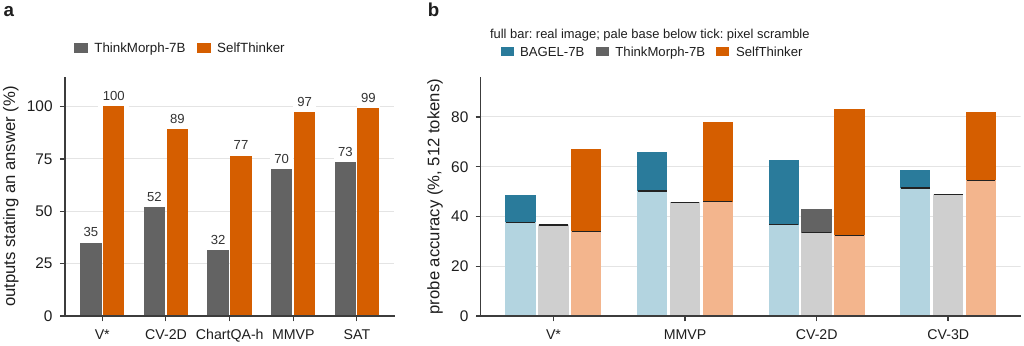}
\caption{Analysis of the distilled student. \textbf{a} Percentage of outputs that state a final answer before the image slot under the benchmark protocol, for ThinkMorph-7B and \method. \textbf{b} Image-condition probe at 512 thinking tokens for pretrained BAGEL-7B, ThinkMorph-7B, and \method across V*, MMVP, CV-Bench-2D, and CV-Bench-3D. Full bar height gives accuracy with the real problem image, and the pale base below the horizontal tick gives accuracy on its pixel scramble, so the upper saturated segment represents the visual gain}\label{fig:why}
\end{figure*}

\paragraph{Cross-task gains on vision-centric reasoning.}
Grouping the benchmarks by task type shows how self-generated visual supervision transfers to diverse visual domains. On fine-grained visual reading and search tasks that require locating small objects or reading dense charts, \emph{\method improves ChartQA by $+27.4$ points (to 80.1, and 81.9 for \method (Vote)) and V* by $+16.3$ points (to 70.2)}. On multi-step spatial planning, \emph{\method lifts VSP maze navigation from near-failure (1.2) to 15.8 ($+14.6$ points) without drawing intermediate paths at test time}, while improving perspective reasoning on SAT by $+4.7$ points (to 60.0). On 2D and 3D natural-scene perception, \emph{\method consistently advances CV-Bench ($+8.3$, to 81.7), BLINK ($+4.6$, to 58.2), and MMVP (\textbf{$+3.3$}, to 82.3)}.

\subsection{Cross-Task Perception Transfer and Internalization}\label{sec:res-why}

\paragraph{Unlocking multi-scale visual perception beyond the pretrained BAGEL-7B backbone.}
To verify that these cross-task gains reflect stronger visual perception rather than language shortcuts, we compare \method against both ThinkMorph-7B and its pretrained BAGEL-7B backbone under the 512-token image-condition probe (Fig.~\ref{fig:why}b; full 256- and 512-token corruption tables in Appendix~\ref{app:probe}). Across all four spatial and object-perception benchmarks, \emph{\method surpasses ThinkMorph-7B by 33.0 to 40.4 points and outperforms the pretrained BAGEL-7B backbone by 12.0 to 23.2 points}. Specifically, \method reaches 67.0\% on V* ($+18.3$ over BAGEL-7B), 78.0\% on MMVP ($+12.0$), 83.4\% on CV-Bench-2D ($+20.9$), and 82.0\% on CV-Bench-3D ($+23.2$), while also lifting ChartQA-h/v under the 256-token probe from 32.2\%--32.5\% in ThinkMorph-7B to 67.3\%--68.6\% (Table~\ref{tab:probe1}). Because pixel-scramble accuracy remains at a similar baseline across all three models, \emph{the entire improvement over BAGEL-7B stems from stronger visual evidence extraction from the real problem image}. Graded mosaic, Gaussian blur, and tile-permutation ladders in Appendix~\ref{app:probe} further show that \method preserves global inter-object geometry across $2\times2$ to $8\times8$ mosaics and mild blur on 2D and 3D spatial tasks, while extracting fine local glyphs on ChartQA.

\paragraph{Eliminating premature termination at the image slot.}
As shown in Fig.~\ref{fig:why}a, when test-time image generation is disabled, ThinkMorph-7B frequently halts prematurely at the \texttt{<image\_start>} token, completing a final text answer on only 32\% to 73\% of items across V*, CV-Bench-2D, ChartQA-h, MMVP, and SAT. \emph{\method raises direct text answer completion to 77\%--100\% (reaching 100\% on V* and 99\% on SAT)}. Crucially, on the paired subsets of items where ThinkMorph-7B already states an answer before \texttt{<image\_start>}, \emph{\method still improves conditional accuracy under the 512-token probe from 35.3\% to 66.8\% on V* ($+31.5$ points), from 46.6\% to 82.9\% on CV-Bench-2D ($+36.3$), from 37.2\% to 68.4\% on ChartQA-h ($+31.3$), from 48.9\% to 77.4\% on MMVP ($+28.5$), and from 50.4\% to 60.0\% on SAT ($+9.6$)}. This confirms that self-distillation strengthens visual reasoning itself rather than merely repairing answer formatting.

\paragraph{Internalizing visual thoughts without test-time drawing.}
To verify that the student has absorbed the privileged teacher conditioning into single-image reasoning, we perform controlled context interventions on the 1,092 verified visual-search items (Table~\ref{tab:intv} and Fig.~\ref{fig:intv} in Appendix~\ref{app:decode}). Whereas ThinkMorph-7B drops by 6.1 points when the verified visual thought is removed at test time (45.9\% with the thought versus 39.7\% with the problem image alone), \emph{\method achieves virtually identical accuracy whether given the problem plus its verified thought (99.0\%) or the problem image alone (98.9\%, a difference of $+0.1$ points)}. Label-free \method (Vote) likewise matches at 95.4\% versus 95.3\% ($+0.1$ points). Conversely, when the problem image is withheld and only the verified thought is shown, \emph{\method and \method (Vote) reach 96.5\% and 93.0\% accuracy (versus 35.4\% for ThinkMorph-7B)}, showing that distillation sharpens the model's visual reader. To evaluate generalization on unseen problem types, we also test \method (Nav), which is distilled solely on grid navigation and never exposed to visual-search items. \emph{\method (Nav) transfers zero-shot to the 1,092 visual-search problems at 92.9\% single-image accuracy (with a $0.0$-point gap between with and without the visual thought, up from 39.7\% in ThinkMorph-7B)}.

\subsection{Intrinsic Verification Analysis}\label{sec:res-audit}

\begin{table*}[t]
\caption{Reliability verification on the diagnostic splits (500 items each). Yield is the fraction of items with a verified thought, Top-1 AUC compares the top intrinsic thought with the mean of mismatched utilities, Win rate is the percentage of intrinsic thoughts that outscore their paired mismatched thought, and Margin is the mean utility difference in nats}\label{tab:audit}
\footnotesize
\setlength{\tabcolsep}{4pt}
\begin{tabular*}{\textwidth}{@{\extracolsep\fill}L{0.36\textwidth}cccc@{}}
\toprule
Task and direction & Yield & Top-1 AUC & Win rate (\%) & Margin (nats) \\
\midrule
Navigation, answer direction & \textbf{0.674} & \textbf{0.821} & \textbf{57.0} & \textbf{2.10} \\
Visual search, answer direction & 0.578 & 0.660 & 54.1 & 0.20 \\
Visual search, vote direction (label-free) & 0.578 & 0.647 & 54.5 & 0.21 \\
\quad accurate-vote subset (93\% of items) & 0.583 & 0.654 & 54.9 & 0.23 \\
\botrule
\end{tabular*}
\end{table*}

\begin{figure*}[t]
\centering
\includegraphics[width=\textwidth]{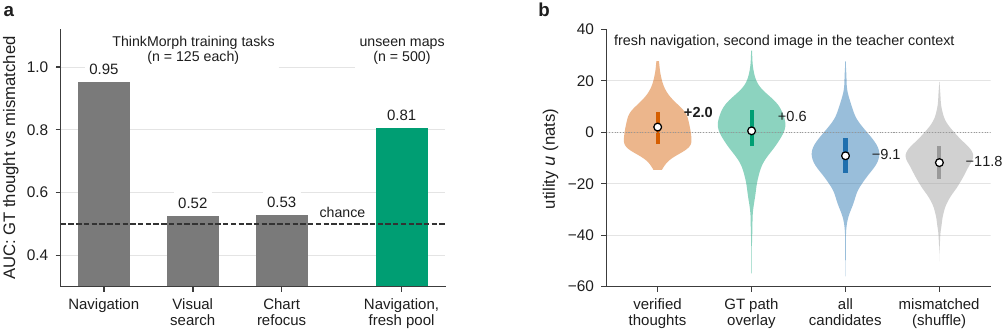}
\caption{What intrinsic verification measures. \textbf{a} Ground-truth thoughts of ThinkMorph's training tasks and of fresh navigation maps, verified against mismatched ground-truth thoughts. \textbf{b} Utility $u$ (nats) on fresh navigation for verified intrinsic thoughts $v^\star$, programmatic GT path overlays, all self-generated candidates, and mismatched thoughts from other maps. Dots show means and vertical bars show interquartile ranges}\label{fig:audit}
\end{figure*}

\begin{figure*}[t]
\centering
\includegraphics[width=\textwidth]{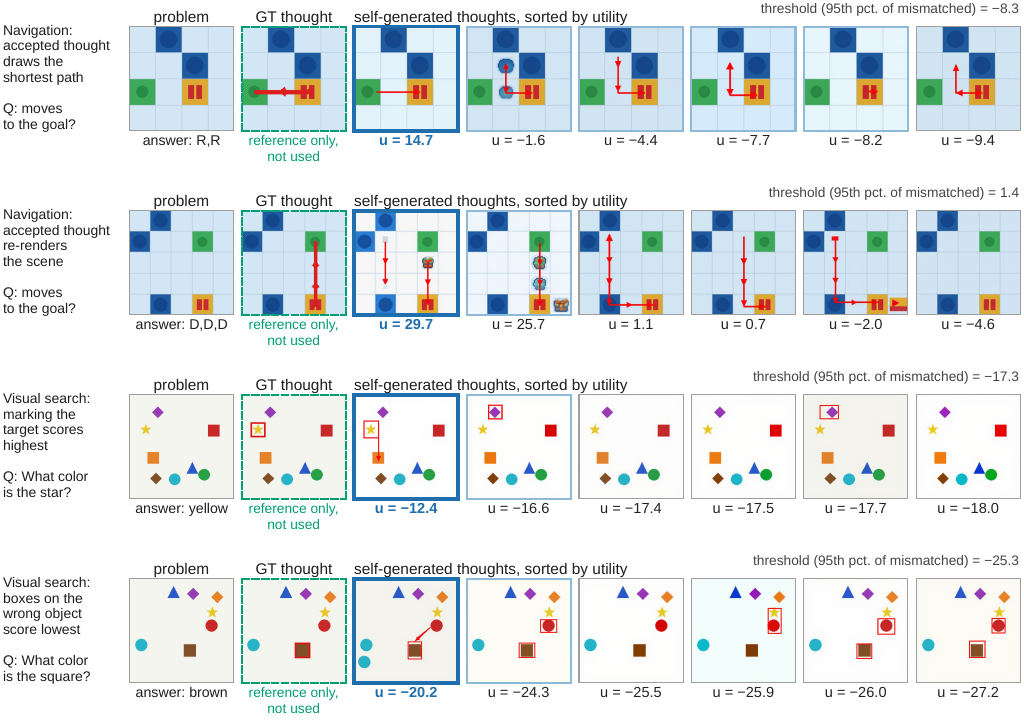}
\caption{Verified visual thoughts on diagnostic items. Each row shows the problem, the programmatic reference thought (dashed green frame, not used by verification), and the six intrinsic visual thoughts sorted by utility $u$ (nats). The dark blue frame marks the verified top thought, and light blue frames mark other thoughts exceeding the item's 95th-percentile mismatched threshold printed above each row. Row 1: the verified thought draws the shortest path. Row 2: the verified thoughts re-render the scene consistently. Rows 3 and 4 (visual search): marking the queried target scores highest, whereas boxes on distractor objects score lowest}\label{fig:examples}
\end{figure*}

\paragraph{Why the understanding pathway must select what is worth learning.}
We first examine how much answer evidence the frozen understanding pathway extracts from ThinkMorph's human-annotated helper images across its training tasks (Fig.~\ref{fig:audit}a). While shortest-path overlays carry strong utility on both training maps (AUC 0.95) and unseen fresh maps (AUC 0.81), fixed human templates on visual search and chart refocus yield near-chance discrimination between matched and mismatched thoughts (AUC 0.52--0.53). \emph{Prescribing a fixed helper-image template per task does not guarantee that the model's reader can extract useful evidence from it; scoring candidate thoughts directly through the understanding pathway provides a task-agnostic criterion for selecting informative visual expressions.}

\paragraph{Separating helpful visual cues from conflicting imagery.}
Table~\ref{tab:audit} and Fig.~\ref{fig:audit}b evaluate intrinsic verification on self-generated thoughts across the unseen diagnostic splits. On navigation, unverified candidate rollouts average $-9.1$ nats and mismatched thoughts from other maps reduce target answer log-likelihood by $-11.8$ nats on average. In contrast, \emph{verified intrinsic thoughts $v^\star$ achieve a positive mean utility of $+2.0$ nats, exceeding programmatic GT path overlays ($+0.6$ nats; Fig.~\ref{fig:audit}b)}, while reaching \emph{a Top-1 AUC of 0.821, a verification yield of 0.674, and a utility margin of 2.10 nats} (Table~\ref{tab:audit}). On visual search, appending a second image introduces a constant baseline log-likelihood offset across candidates (rows 3--4 of Fig.~\ref{fig:examples}). Calibrating against same-item mismatched thoughts $u_i(\vmis^{(k)})$ removes this offset and achieves \emph{a Top-1 AUC of 0.660 and a yield of 0.578}. Comparing generated red boxes against the ground-truth target coordinates on the 500 unseen visual-search diagnostic items further confirms that \emph{thoughts boxing the true queried target achieve a $+0.31$-nat utility margin over mismatched thoughts, whereas boxes placed on distractor objects average $-0.04$ nats}.

\paragraph{Label-free verification via rollout majority voting.}
We next test whether visual thoughts can be verified without ground-truth answer labels. On visual search, 47.4\% of diagnostic items exhibit high text answer agreement across visibly diverse visual thoughts, and trace answer correctness is nearly orthogonal to the visual utility of the accompanying thought ($\rho=0.022$; full analysis in Appendix~\ref{app:auditstats}). Consequently, text consensus alone cannot certify image quality. However, when the rollout majority vote $\tilde a$ is used to set the target direction $d$ in Eq.~\eqref{eq:utility}, \emph{label-free vote verification closely matches ground-truth answer verification (Table~\ref{tab:audit}), achieving a 0.895 Spearman rank correlation with ground-truth utilities, a 54.5\% paired win rate, and a 63.4 seven-benchmark average ($+10.7$ points over ThinkMorph-7B, Table~\ref{tab:bench})}.

\subsection{Ablation Studies}\label{sec:res-abl}

\begin{table*}[t]
\caption{Ablation of the self-evolving understanding update across training task pools, teacher contexts, and supervision directions. Average gives the seven-benchmark mean and $\Delta$ gives the improvement over ThinkMorph-7B}\label{tab:ablund}
\footnotesize
\setlength{\tabcolsep}{4pt}
\begin{tabular*}{\textwidth}{@{\extracolsep\fill}L{0.18\textwidth}L{0.49\textwidth}cc@{}}
\toprule
Method & Configuration & Average & $\Delta$ vs.\ ThinkMorph-7B \\
\midrule
ThinkMorph-7B & Base interleaved model & 52.7 & -- \\
\midrule
\method (Nav) & On-policy self-distillation with verified navigation thoughts & 60.0 & $+7.3$ \\
\method (Mix) & On-policy self-distillation with combined navigation and visual-search thoughts & 60.8 & $+8.1$ \\
Prompt-OPSD & Visual-search distillation with transition prompt only (no second image) & 60.7 & $+8.0$ \\
Noise-OPSD & Visual-search distillation with uniform RGB noise in place of the thought & 61.8 & $+9.1$ \\
\method (Vote) & Visual-search distillation with thoughts verified via majority voting (label-free) & 63.4 & $+10.7$ \\
\textbf{\method} & Visual-search distillation with verified visual thoughts & \textbf{64.0} & \textbf{$+11.3$} \\
\botrule
\end{tabular*}
\end{table*}

\begin{table*}[t]
\caption{Ablation of the supporting generation update on the diagnostic splits (500 items per task). Each generator draws six visual thoughts per item, and the frozen ThinkMorph-7B understanding pathway scores them. \method-Gen improves every verification metric over both ThinkMorph-7B and Rand-RFT}\label{tab:ablgen}
\footnotesize
\setlength{\tabcolsep}{4pt}
\begin{tabular*}{\textwidth}{@{\extracolsep\fill}llcccc@{}}
\toprule
Task & Method & Margin (nats) & Win rate (\%) & Top-1 AUC & Yield \\
\midrule
Navigation & ThinkMorph-7B & 2.10 & 57.0 & 0.821 & 0.674 \\
Navigation & Rand-RFT & 2.06 & 57.3 & 0.825 & 0.659 \\
Navigation & \method-Gen & \textbf{2.96} & \textbf{60.0} & \textbf{0.847} & \textbf{0.694} \\
\midrule
Visual search & ThinkMorph-7B & 0.20 & 54.1 & 0.660 & 0.578 \\
Visual search & \method-Gen & \textbf{0.27} & \textbf{54.4} & \textbf{0.675} & \textbf{0.632} \\
\botrule
\end{tabular*}
\end{table*}

\paragraph{Privileged visual thoughts versus context distillation.}
Table~\ref{tab:ablund} disentangles the factors driving \method's benchmark gains (complete per-benchmark breakdown in Appendix~\ref{app:full}). First, distilling verified thoughts into the understanding pathway via on-policy self-distillation yields large gains across all task pools. Performance scales consistently from navigation (\method (Nav) at 60.0, $+7.3$) and mixed pools (\method (Mix) at 60.8, $+8.1$) to visual search (\method at 64.0, $+11.3$, and label-free \method (Vote) at 63.4, $+10.7$). Across 13 independent training seeds for \method ($62.4 \pm 1.0$, range $60.7$--$64.0$) and 11 seeds for label-free \method (Vote) ($61.9 \pm 0.8$, range $61.1$--$63.4$), every run matches or outperforms Prompt-OPSD (60.7). Second, paired item-level bootstrap tests (10,000 resamples) confirm that \emph{conditioning the teacher on verified visual thoughts (\method at 64.0) significantly outperforms both Prompt-OPSD ($+3.3$ points, 95\% CI $[+1.7, +5.0]$, $p < 10^{-4}$) and Noise-OPSD ($+2.2$ points, 95\% CI $[+0.8, +3.8]$, $p = 0.0015$)}. Label-free \method (Vote) (63.4) likewise significantly exceeds both controls ($+2.7$ points, CI $[+1.0, +4.2]$; $+1.6$ points, CI $[+0.1, +3.1]$), while verified visual thoughts further provide $+10.7$ to $+23.3$ point gains on fine-grained 2D and 3D image-condition probes (Appendix~\ref{app:probe}).

\paragraph{Supporting generative refinement improves candidate training imagery.}
Table~\ref{tab:ablgen} evaluates the supporting generative update (\method-Gen) on candidate thought generation across the 500-item unseen diagnostic splits. Training on randomly selected thoughts (\emph{Rand-RFT}) fails to improve utility margins on unseen navigation maps (2.06 versus 2.10 nats). In contrast, \emph{\method-Gen raises the navigation utility margin to 2.96 nats ($+0.90$ nats over Rand-RFT), lifts Top-1 AUC from 0.821 to 0.847, and increases yield from 0.674 to 0.694}. On visual search, \emph{\method-Gen improves every verification metric (lifting yield from 0.578 to 0.632 and Top-1 AUC from 0.660 to 0.675) while increasing the localized bounding-box generation rate from 40.8\% to 50.8\%} (Appendix~\ref{app:auditstats}), providing cleaner visual supervision for understanding evolution.

\section{Conclusion}\label{sec:conclusion}

We introduced \method, a self-evolving framework that turns a unified multimodal model's image generation capability into training supervision for visual understanding without human image annotations. By scoring candidate visual thoughts through the model's frozen understanding pathway via calibrated conditional pointwise mutual information, \method selects informative helper images under both ground-truth answers and label-free majority votes. Distilling these verified visual thoughts into the understanding pathway via privileged on-policy self-distillation, supported by flow-matching generator refinement, enables the deployed model to reason directly from the original image in text without generating images at test time. Although trained only on small synthetic visual-search and navigation pools, \method improves the seven-benchmark average of ThinkMorph-7B by $+11.3$ points (and by $+10.7$ points under label-free majority voting). Controlled image-perturbation probes further show that this self-generated visual supervision transfers broadly across tasks, unlocking fine-grained and 2D/3D spatial perception that surpasses ThinkMorph-7B by up to 40.4 points and the pretrained BAGEL-7B backbone by up to 23.2 points.

% \input{Sec/declarations}
% \clearpage
% \newpage
\begin{appendices}
\renewcommand{\thetable}{\arabic{table}}
\renewcommand{\theHtable}{app.\arabic{table}}
\setcounter{table}{5}
\renewcommand{\thefigure}{\arabic{figure}}
\renewcommand{\theHfigure}{app.\arabic{figure}}
\setcounter{figure}{5}

\section{Implementation Details}\label{app:impl}

\begin{table*}[t]
\caption{Synthesized problem pools built outside the training data of ThinkMorph-7B. Each pool contains 1,500 training items and 500 diagnostic items. Programmatic reference thoughts are used only as diagnostic baselines and never enter verification or model training}\label{tab:pools}
\footnotesize
\begin{tabular*}{\textwidth}{@{\extracolsep\fill}L{0.12\textwidth}L{0.32\textwidth}L{0.30\textwidth}L{0.14\textwidth}@{}}
\toprule
Pool & Problem image & Question and answer & Reference thought \\
\midrule
Navigation & Frozen-Lake grid, $4\times4$ to $8\times8$, 12--28\% holes, $192\times192$ px & Move sequence along the unique shortest path (2--14 moves, mean 5.1) & Path overlay \\
Visual search & 5--8 non-overlapping shapes (5 shapes, 8 colours), $384\times384$ px & Colour of a target shape or shape of a target colour (13 classes) & Red bounding box \\
\botrule
\end{tabular*}
\end{table*}

\begin{table*}[t]
\caption{Training hyperparameters for \method-Gen and \method. Both updates start from ThinkMorph-7B and run on one node with eight NVIDIA H20 GPUs}\label{tab:hparams}
\footnotesize
\begin{tabular*}{\textwidth}{@{\extracolsep\fill}L{0.18\textwidth}L{0.38\textwidth}L{0.38\textwidth}@{}}
\toprule
Setting & \method-Gen (generation update) & \method (understanding update) \\
\midrule
Trainable pathway & Generation and understanding (shared decoder) & Understanding; generation pathway unused \\
Objective & Flow matching on thought latents (weight 1.0), cross-entropy on answer (0.25) & Token-level JSD, $\beta=0.5$, temperature 1.0, top-$k$ 256, per-token clip 0.05 \\
Teacher & -- & EMA of the student, decay 0.995 \\
Student sampling & -- & Temperature 1.0, up to 128 new tokens, one image skip \\
Schedule & 1,000 steps; cosine decay, warmup 100, learning rate $10^{-5}$ to $10^{-7}$ & 1,000 steps; cosine decay, warmup 200, learning rate $10^{-5}$ to $10^{-7}$ \\
Optimizer & AdamW & AdamW $(0.9, 0.95)$, gradient clipping 1.0 \\
Sequence budget & 16,384 tokens per packed batch (8,192 expected) & 10,240 tokens per forward pass \\
Parallelism & Fully sharded data parallel over 8 GPUs & Fully sharded data parallel over 8 GPUs \\
Image transforms & Vision Transformer (ViT) stride 14, 378--980 px; variational autoencoder (VAE) stride 16, 512--1,024 px & ViT stride 14, 378--980 px, at most 2,007,040 pixels \\
\botrule
\end{tabular*}
\end{table*}

\paragraph{Rollout sampling.}
Table~\ref{tab:pools} summarizes the synthesized navigation and visual-search problem pools, and Table~\ref{tab:hparams} lists the training hyperparameters for \method-Gen and \method. For each item, we sample $K=6$ interleaved traces with one visual thought each using the interleaved inferencer of ThinkMorph-7B. Sampling uses 10 denoising steps per thought, text temperature 0.7, classifier-free guidance scales of 4.0 for text and 2.0 for images, and at most 256 thinking tokens. Generating one interleaved trace takes 59\,s on navigation and 75\,s on visual search on one NVIDIA H20 GPU, whereas scoring a candidate thought with its controls takes 2.3--2.8\,s. Navigation thoughts are sampled from ThinkMorph-7B (and from each fine-tuned generator for generator comparisons), while visual-search thoughts are sampled from the navigation \method-Gen checkpoint. All understanding students are initialized from ThinkMorph-7B.

\paragraph{Threshold calibration.}
For intrinsic verification in Eq.~\eqref{eq:utility}, the plain context $C(I,q)$ and privileged context $C(I,q,\vt)=[I,q,\rho,\vt,\tau]$ are formatted identically to the student and teacher contexts of Eq.~\eqref{eq:opsd}. We use the transition instruction $\tau$ (``Now, using your own independent reasoning, answer the problem above. Think step by step.'') and the privileged context header $\rho$ (``The following images are additional visual context. Use them silently as grounding context, and do not describe or echo them.''). The mismatched partner $\vmis^{(k)}$ of each thought is assigned by a fixed cyclic offset across items in the scoring order. This ensures that every candidate is compared against an in-distribution thought generated by the same model for a distinct problem. For training set construction, the 95th-percentile threshold $Q_i$ uses the order-statistic form, i.e., the $(\lfloor 0.95(K-1)\rfloor+1)$-th smallest element of the mismatched utilities $s_i$ (the second largest for $K=6$), whereas diagnostic yield calculations use the linearly interpolated 95th percentile.

\paragraph{Teacher-context controls.}
Navigation training uses 1,179 verified thoughts from 1,472 scored training items. Visual search uses 1,092 verified thoughts (or 1,074 under the vote direction) from 1,500 items, and the mixed pool combines both verified sets. In \method-Gen, the flow-matching loss on latent thought tokens has weight 1.0 and the cross-entropy loss on answer tokens has weight 0.25. Rand-RFT uses the exact same 1,179 navigation items and optimization schedule as \method-Gen, but selects one thought per item uniformly at random without reading $u$. The teacher-context controls on visual search (Noise-OPSD and Prompt-OPSD) reuse the exact items and optimization settings in Table~\ref{tab:hparams} (1,000 steps on eight NVIDIA H20 GPUs with learning rate $10^{-5}$ and cosine decay, with navigation distillation trained for 2,000 steps). They modify only the teacher context $c_T$: replacing $\vt^\star$ with a $512\times512$ image of independent uniform RGB noise in Noise-OPSD, or omitting $\vt^\star$ in Prompt-OPSD.

\begin{table*}[t]
\caption{Per-benchmark breakdown of task-pool variants, teacher-context controls, and \method under the unified judge and text-only decoding. $\Delta$ gives the improvement in the seven-benchmark average relative to ThinkMorph-7B. CV-B.: CV-Bench}\label{tab:bench}
\footnotesize
\setlength{\tabcolsep}{4.6pt}
\begin{tabular*}{\textwidth}{@{\extracolsep\fill}L{0.22\textwidth}cccccccc r@{}}
\toprule
Method & VSP & ChartQA & V* & MMVP & SAT & BLINK & CV-B. & Average & $\Delta$ \\
\midrule
ThinkMorph-7B & 1.2 & 52.7 & 53.9 & 79.0 & 55.3 & 53.6 & 73.4 & 52.7 & -- \\
\midrule
\multicolumn{10}{@{}l}{\emph{Understanding evolution: navigation and mixed pools}} \\
\method (Nav) & 2.0 & 78.1 & 63.4 & 80.3 & 58.0 & \textbf{60.3} & 78.0 & 60.0 & $+7.3$ \\
\method (Mix) & 5.5 & 74.7 & \textbf{70.2} & 79.3 & 54.7 & 58.0 & \textbf{83.5} & 60.8 & $+8.1$ \\
\midrule
\multicolumn{10}{@{}l}{\emph{Understanding evolution: visual-search pool and teacher controls}} \\
Prompt-OPSD & 3.3 & 77.7 & 68.6 & 79.7 & 56.7 & 58.4 & 80.6 & 60.7 & $+8.0$ \\
Noise-OPSD & 13.2 & 78.3 & 65.4 & 80.0 & 56.0 & 58.5 & 81.2 & 61.8 & $+9.1$ \\
\method (Vote, label-free) & 13.0 & \textbf{81.9} & 67.0 & 80.0 & \textbf{60.0} & 59.3 & 82.4 & 63.4 & $+10.7$ \\
\textbf{\method} & \textbf{15.8} & 80.1 & \textbf{70.2} & \textbf{82.3} & \textbf{60.0} & 58.2 & 81.7 & \textbf{64.0} & \textbf{$+11.3$} \\
\textbf{$\Delta$ vs.\ ThinkMorph-7B} & \textbf{$+14.6$} & \textbf{$+27.4$} & \textbf{$+16.3$} & \textbf{$+3.3$} & \textbf{$+4.7$} & \textbf{$+4.6$} & \textbf{$+8.3$} & \textbf{$+11.3$} & -- \\
\botrule
\end{tabular*}
\end{table*}

\begin{figure*}[t]
\centering
\includegraphics[width=\textwidth]{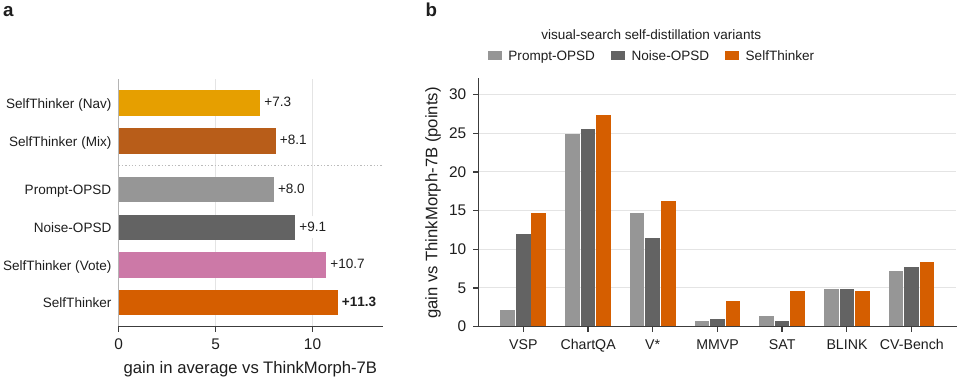}
\caption{Comparison of \method variants and teacher-context controls. \textbf{a} Gain in the seven-benchmark average relative to ThinkMorph-7B across navigation, mixed-pool, and visual-search distillation runs. \textbf{b} Per-benchmark gains relative to ThinkMorph-7B for Prompt-OPSD, Noise-OPSD, and \method}\label{fig:teach}
\end{figure*}

\section{Diagnostic Definitions}\label{app:defs}
\paragraph{Verification metrics.}
On the 500-item diagnostic splits, the \emph{utility margin} is the mean difference $u(\vt)-u(\vmis)$ in nats across all candidate pairs, and the \emph{win rate} is the percentage of pairs with $u(\vt)>u(\vmis)$. The \emph{Top-1 AUC} computes the area under the ROC curve comparing each item's top intrinsic thought against the mean of its mismatched utilities \citep{hanley1982auc}, and the \emph{yield} is the fraction of items whose top thought exceeds $Q_i$. As an analytical diagnostic for label-free verification, we also report metrics on the accurate-vote subset where the rollout majority vote $\tilde a$ matches $a^\star$.
\paragraph{Bounding-box detector.} Each visual-search thought is converted to RGB, and a pixel is classified as red if $r\geq160$, $g\leq90$, $b\leq90$, and $r-\max(g,b)>50$. A thought is flagged as containing a red bounding box if its red pixel fraction lies in $[0.0015, 0.18]$ with at least 30 red pixels, its red bounding box is at least $10\times10$ pixels, and the red density along the perimeter band of the bounding box is at least 0.08 and exceeds the interior red density by at least 0.04.
\paragraph{Dual consensus.} On visual search, an item has high answer agreement if the majority vote accounts for at least 80\% of the $K=6$ rollout answers. It has high visual agreement if the mean pairwise cosine similarity of its six thoughts (after resizing to $48\times48$ RGB arrays and subtracting each array's mean) exceeds the median across items.
\paragraph{Pixel deviation.} Pixel deviation measures the mean absolute difference between a visual thought and its problem image after resizing both to $96\times96$ RGB in $[0,1]$. Programmatic path overlays deviate by 0.015 on average, and self-generated navigation thoughts have quartile boundaries of 0.012, 0.017, and 0.043.

\begin{figure*}[t]
\centering
\includegraphics[width=\textwidth]{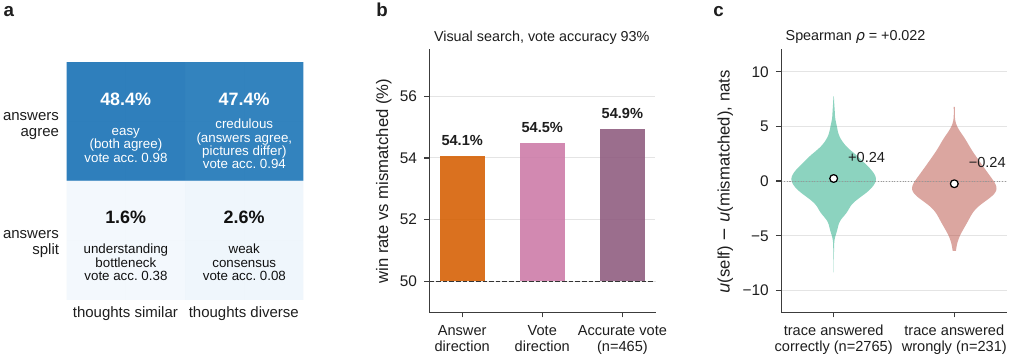}
\caption{Answer consensus versus imagery on visual search (500 diagnostic items). \textbf{a} Items split by answer agreement (vote share at least 0.8) and visual agreement (mean pairwise cosine similarity of the six thoughts, median split), with the vote accuracy of each cell. \textbf{b} Paired win rate against mismatched thoughts under the answer direction, the vote direction, and the accurate-vote subset. \textbf{c} Utility difference $u(\text{self})-u(\text{mismatched})$ for thoughts whose trace answered correctly versus wrongly}\label{fig:consensus}
\end{figure*}

\begin{figure*}[t]
\centering
\includegraphics[width=\textwidth]{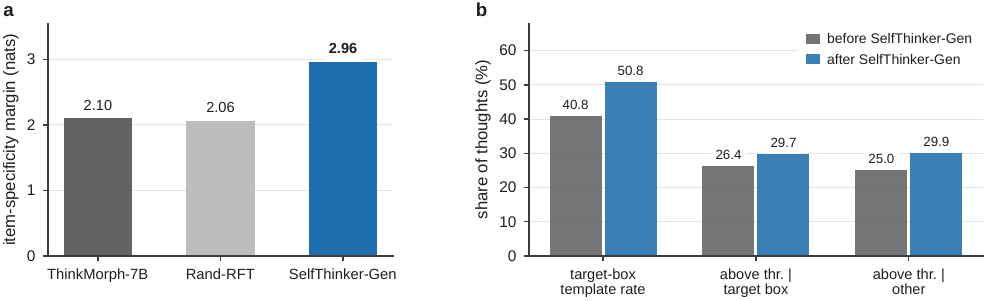}
\caption{Improvements in the generation pathway under \method-Gen. \textbf{a} Utility margin (nats) of navigation thoughts drawn by ThinkMorph-7B, Rand-RFT, and \method-Gen on the 500-item diagnostic split. \textbf{b} Bounding-box generation rate and above-threshold pass rates on visual search before and after \method-Gen}\label{fig:gen}
\end{figure*}

\section{Per-Benchmark Breakdown of Distillation Variants}\label{app:full}

Table~\ref{tab:bench} reports the complete per-benchmark scores of all task-pool variants, visual-search teacher controls, and \method. Figure~\ref{fig:teach} visualizes the overall and per-benchmark gains relative to ThinkMorph-7B.

Across training pools (Fig.~\ref{fig:teach}a), navigation distillation in \method (Nav) lifts the seven-benchmark average by $+7.3$ points (reaching 60.0 and achieving the highest score of 60.3 on BLINK). Combining navigation with visual-search thoughts in \method (Mix) raises the gain to $+8.1$ points (60.8) and achieves the highest score on CV-Bench (83.5). Moving to the visual-search pool yields a further jump: Prompt-OPSD ($+8.0$, 60.7) and Noise-OPSD ($+9.1$, 61.8) establish strong context-distillation baselines, and \emph{conditioning the teacher on verified visual thoughts raises the gain to $+10.7$ points for label-free \method (Vote) (63.4, reaching 81.9 on ChartQA) and $+11.3$ points for \method (64.0)}. Across individual benchmarks (Fig.~\ref{fig:teach}b), \method consistently improves over Prompt-OPSD and Noise-OPSD on ChartQA (80.1 versus 77.7 and 78.3), V* (70.2 versus 68.6 and 65.4), VSP (15.8 versus 3.3 for Prompt-OPSD and 13.2 for Noise-OPSD), MMVP (82.3 versus 79.7 and 80.0), SAT (60.0 versus 56.7 and 56.0), and CV-Bench (81.7 versus 80.6 and 81.2).

\begin{table*}[t]
\caption{Context interventions on the 1,092 verified visual-search items (accuracy in \%). Own: the item's verified visual thought as second image. None: no second image (single-image deployment). Neutral: a neutral sentence appended. True ans.: the ground-truth answer appended. Thought only: the verified visual thought replaces the problem image. The last column reports the difference between Own and None, measuring single-image internalization}\label{tab:intv}
\footnotesize
\setlength{\tabcolsep}{4.5pt}
\begin{tabular*}{\textwidth}{@{\extracolsep\fill}L{0.22\textwidth}rrrrrr@{}}
\toprule
Method & Own & None & Neutral & True ans. & Thought only & Own $-$ None \\
\midrule
ThinkMorph-7B & 45.9 & 39.7 & 38.0 & 40.7 & 35.4 & $+6.1$ \\
\midrule
\method (Nav) & 92.9 & 92.9 & 91.2 & 96.4 & 89.1 & $+0.0$ \\
\method (Vote) & 95.4 & 95.3 & 95.3 & 99.7 & 93.0 & $+0.1$ \\
\textbf{\method} & 99.0 & 98.9 & 98.9 & 99.5 & 96.5 & $+0.1$ \\
\botrule
\end{tabular*}
\end{table*}

\begin{figure*}[t]
\centering
\includegraphics[width=\textwidth]{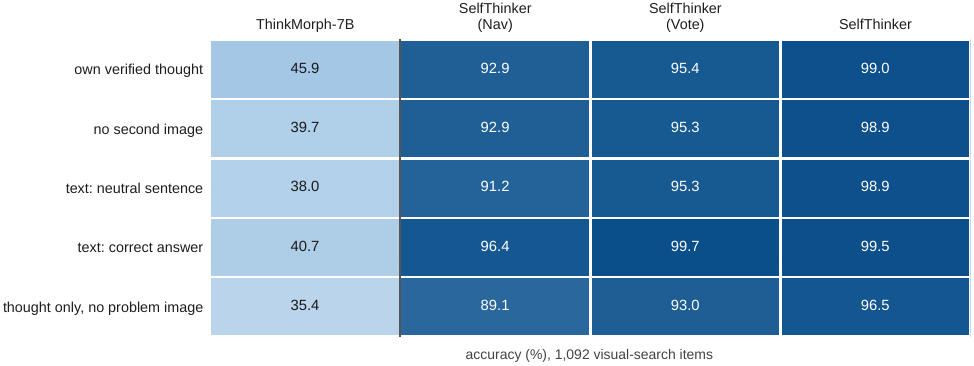}
\caption{Context interventions on the 1,092 verified visual-search items across ThinkMorph-7B, \method (Nav), \method (Vote), and \method under five conditions (accuracy in \%). Rows compare providing the item's verified thought as second image, no second image (single-image deployment), appending a neutral sentence or the ground-truth answer, or replacing the problem image with the verified thought alone}\label{fig:intv}
\end{figure*}

\section{Additional Verification Analyses}\label{app:auditstats}

\paragraph{Acceptance across pixel deviation quartiles.}
On navigation, \emph{verified thoughts span the full range of pixel deviation from the problem map}, with acceptance rates of 11.4\%, 13.9\%, 11.6\%, and 15.1\% across the four deviation quartiles. The highest-deviation quartile, which includes complete scene re-renderings such as row 2 of Fig.~\ref{fig:examples}, achieves the highest acceptance rate (15.1\%).

\paragraph{Answer consensus versus visual consensus.}
Figure~\ref{fig:consensus}a shows the dual-consensus breakdown across the 500 diagnostic visual-search items: 48.4\% of items have both high answer agreement and visually similar thoughts, while 47.4\% have high answer agreement produced from visibly diverse thoughts (matching the 48.5\% and 47.3\% split on the 1,500 training items). In this diverse-imagery regime, \emph{intrinsic verification identifies a validated helper image on 66\% of items}. Figure~\ref{fig:consensus}b,c further shows that \emph{the label-free vote direction matches the answer direction in paired win rate (54.5\% versus 54.1\%, and 54.9\% on accurate-vote items)}, while trace answer correctness is orthogonal to the visual utility of the thought ($\rho=+0.022$).

\paragraph{Generation pathway gains and target-box selectivity.}
Figure~\ref{fig:gen} illustrates how \method-Gen improves the generation pathway on both tasks. On navigation (Fig.~\ref{fig:gen}a), \emph{\method-Gen lifts the utility margin on unseen diagnostic maps from 2.10 nats (ThinkMorph-7B) and 2.06 nats (Rand-RFT) to 2.96 nats}, confirming that verification drives the improvement in generated imagery. On visual search (Fig.~\ref{fig:gen}b), \emph{\method-Gen increases the share of thoughts that draw a localized red bounding box from 40.8\% to 50.8\%, while raising the utility-threshold pass rate among both box thoughts (from 26.4\% to 29.7\%) and non-box thoughts (from 25.0\% to 29.9\%)}. Evaluating against the ground-truth target coordinates on the 500 unseen diagnostic items further confirms that \emph{thoughts boxing the true queried object achieve a mean utility margin of $+0.31$ nats over mismatched thoughts, compared with $-0.04$ nats for boxes on distractor objects}.

\begin{table*}[t]
\caption{Image-condition probe on ChartQA and V*. Accuracy (\%) under each condition of the problem image (256 thinking tokens). real$^\dagger$ and wrong$^\dagger$: real image and another item's image on the subset whose previous row shows a different image. scr.: pixels in random order. $\times k$: mosaic with $k$-pixel blocks. $\sigma k$: Gaussian blur with $\sigma=k$ pixels. Dashes: not run}\label{tab:probe1}
\footnotesize
\renewcommand{\arraystretch}{0.88}
\setlength{\aboverulesep}{1.2pt}
\setlength{\belowrulesep}{1.2pt}
\setlength{\tabcolsep}{2.5pt}
\begin{tabular*}{\textwidth}{@{\extracolsep\fill}lrrrrrrrrrrrrr@{}}
\toprule
Method & real & real$^\dagger$ & wrong$^\dagger$ & gray & scr. & $\times2$ & $\times4$ & $\times8$ & $\times16$ & $\sigma1$ & $\sigma2$ & $\sigma4$ & $\sigma8$ \\
\midrule
\multicolumn{14}{@{}l}{\emph{ChartQA, horizontal bars (444 items, 222 with a different previous image)}} \\
ThinkMorph-7B & 32.2 & 13.1 & 8.6 & 5.0 & 5.6 & 30.6 & 10.1 & 4.7 & 3.8 & 32.7 & 20.5 & 4.1 & 4.7 \\
\method (Nav) & 51.4 & 33.3 & 11.7 & 9.7 & 10.1 & 48.9 & 19.4 & -- & -- & 48.2 & 36.7 & 10.8 & 10.1 \\
Prompt-OPSD & 63.3 & 52.7 & 9.0 & 11.7 & 10.4 & 59.0 & 24.5 & 12.4 & -- & 60.1 & 45.0 & 14.4 & -- \\
Noise-OPSD & 66.2 & 58.1 & 14.4 & 12.6 & 10.8 & 62.4 & 26.6 & 12.6 & -- & 67.1 & 46.8 & 18.2 & 14.4 \\
\textbf{\method} & 67.3 & 52.3 & 14.9 & 10.4 & 11.3 & 64.9 & 25.7 & 11.7 & 13.7 & 68.9 & 52.3 & 18.2 & 12.8 \\
\midrule
\multicolumn{14}{@{}l}{\emph{ChartQA, vertical bars (382 items, 191 with a different previous image)}} \\
ThinkMorph-7B & 32.5 & 14.1 & 4.2 & 3.4 & 6.0 & 26.7 & 8.6 & 3.7 & 3.9 & 26.4 & 18.3 & 6.5 & 4.5 \\
\method (Nav) & 42.9 & 30.4 & 4.2 & 7.9 & 8.4 & 40.8 & 16.2 & -- & -- & 42.7 & 29.3 & 14.1 & 7.9 \\
Prompt-OPSD & 53.4 & 41.9 & 6.8 & 11.0 & 10.7 & 46.6 & 18.3 & 11.0 & -- & 52.4 & 33.0 & 15.7 & -- \\
Noise-OPSD & 61.0 & 52.9 & 5.8 & 8.6 & 10.2 & 54.7 & 18.1 & 11.3 & -- & 60.2 & 39.5 & 15.7 & 11.8 \\
\textbf{\method} & 68.6 & 54.5 & 4.7 & 11.5 & 13.1 & 63.1 & 19.9 & 12.8 & 11.8 & 63.4 & 43.2 & 15.4 & 14.7 \\
\midrule
\multicolumn{14}{@{}l}{\emph{V* (191 items)}} \\
ThinkMorph-7B & 34.0 & 34.0 & 34.0 & 32.5 & 36.6 & 35.6 & 37.2 & 34.6 & 36.1 & 36.1 & 36.1 & 35.1 & 35.6 \\
\method (Nav) & 56.0 & 56.0 & 40.3 & 38.2 & 38.2 & 57.6 & 50.3 & 52.4 & 38.2 & 59.2 & 56.5 & 53.9 & 45.0 \\
Prompt-OPSD & 50.8 & 50.8 & 36.6 & 34.6 & 36.1 & 51.3 & 49.7 & 45.5 & -- & 52.4 & 51.8 & 47.1 & 41.4 \\
Noise-OPSD & 51.3 & 51.3 & 31.4 & 35.6 & 34.0 & 51.8 & 50.3 & 49.7 & -- & 49.2 & 49.7 & 51.3 & 35.6 \\
\textbf{\method} & 67.0 & 67.0 & 38.2 & 35.1 & 33.0 & -- & 63.9 & 61.8 & 46.1 & 64.4 & 66.5 & 56.5 & 46.1 \\
\botrule
\end{tabular*}
\end{table*}

\begin{figure*}[t]
\centering
\includegraphics[width=\textwidth]{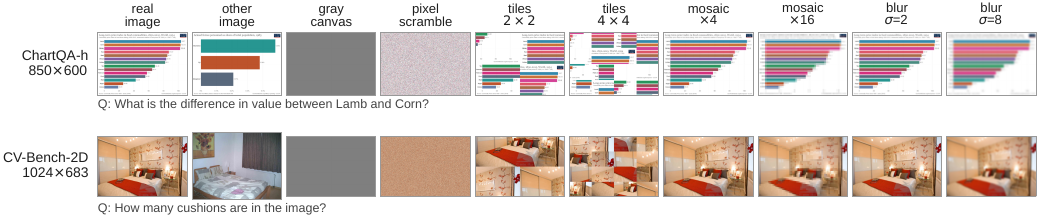}
\caption{Conditions of the image-condition probe, illustrated on a ChartQA horizontal-bar item and a CV-Bench-2D item. The problem image is replaced by another item's image, a gray canvas, its pixels in random order, a $2\times2$ or $4\times4$ tile permutation, a mosaic with square blocks of 4 or 16 pixels, or a Gaussian blur with $\sigma$ of 2 or 8 pixels at the native resolution shown under each benchmark name}\label{fig:probe-examples}
\end{figure*}

\begin{figure*}[t]
\centering
\includegraphics[width=\textwidth]{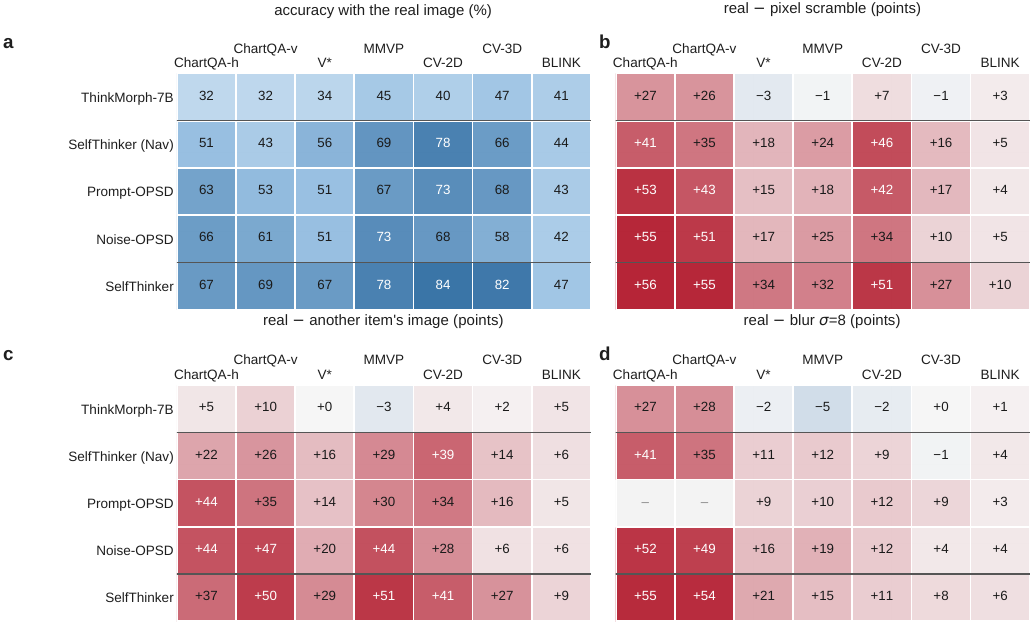}
\caption{Image-condition probe at 256 thinking tokens across ThinkMorph-7B, \method (Nav), Prompt-OPSD, Noise-OPSD, and \method. \textbf{a} Accuracy with the real problem image. \textbf{b}--\textbf{d} Visual sensitivity (accuracy gain in percentage points of the real problem image over pixel scramble, another item's distinct image, and Gaussian blur with $\sigma=8$ pixels)}\label{fig:probe-heat}
\end{figure*}

\begin{table*}[t]
\caption{Image-condition probe on MMVP and CV-Bench. Accuracy (\%) under each condition of the problem image (256 thinking tokens). real$^\dagger$ and wrong$^\dagger$: real image and another item's image on the subset whose previous row shows a different image. scr.: pixels in random order. $\times k$: mosaic with $k$-pixel blocks. $\sigma k$: Gaussian blur with $\sigma=k$ pixels. Dashes: not run}\label{tab:probe2}
\footnotesize
\renewcommand{\arraystretch}{0.88}
\setlength{\aboverulesep}{1.2pt}
\setlength{\belowrulesep}{1.2pt}
\setlength{\tabcolsep}{2.5pt}
\begin{tabular*}{\textwidth}{@{\extracolsep\fill}lrrrrrrrrrrrrr@{}}
\toprule
Method & real & real$^\dagger$ & wrong$^\dagger$ & gray & scr. & $\times2$ & $\times4$ & $\times8$ & $\times16$ & $\sigma1$ & $\sigma2$ & $\sigma4$ & $\sigma8$ \\
\midrule
\multicolumn{14}{@{}l}{\emph{MMVP (300 items)}} \\
ThinkMorph-7B & 44.7 & 44.7 & 47.7 & 49.0 & 45.3 & 45.3 & 47.3 & 49.7 & 47.0 & 47.3 & 51.7 & 46.3 & 50.0 \\
\method (Nav) & 69.3 & 69.3 & 40.0 & 48.3 & 45.0 & 69.7 & 61.7 & 53.0 & 49.7 & 67.7 & 68.7 & 63.3 & 57.3 \\
Prompt-OPSD & 67.0 & 67.0 & 36.7 & 49.7 & 49.0 & 66.7 & 61.0 & 54.0 & -- & 68.0 & 65.0 & 62.3 & 56.7 \\
Noise-OPSD & 73.3 & 73.3 & 29.0 & 49.3 & 48.7 & 70.3 & 61.3 & 56.7 & -- & 72.7 & 69.3 & 62.7 & 54.7 \\
\textbf{\method} & 78.0 & 78.0 & 26.7 & 48.7 & 46.3 & 74.0 & 62.7 & 53.3 & 52.7 & 79.7 & 72.3 & 63.0 & 62.7 \\
\midrule
\multicolumn{14}{@{}l}{\emph{CV-Bench-2D (1,438 items)}} \\
ThinkMorph-7B & 40.2 & 40.2 & 36.4 & 32.6 & 33.4 & 41.5 & 42.6 & 41.9 & 41.0 & 41.2 & 43.1 & 42.9 & 42.4 \\
\method (Nav) & 78.1 & 78.1 & 39.4 & 32.3 & 32.4 & 76.7 & 74.6 & 70.8 & 61.1 & 79.1 & 78.0 & 75.2 & 69.0 \\
Prompt-OPSD & 72.8 & 72.8 & 39.1 & 31.8 & 30.7 & 71.1 & 69.1 & 63.4 & -- & 71.8 & 72.3 & 66.2 & 61.1 \\
Noise-OPSD & 68.3 & 68.3 & 40.5 & 31.6 & 33.8 & 68.6 & 66.1 & 55.4 & -- & 67.6 & 65.6 & 62.1 & 56.3 \\
\textbf{\method} & 83.5 & 83.5 & 42.4 & 33.9 & 32.4 & 82.5 & 80.9 & 75.0 & 67.7 & 84.1 & 81.2 & 78.4 & 72.6 \\
\midrule
\multicolumn{14}{@{}l}{\emph{CV-Bench-3D (1,200 items, 1198 with a different previous image)}} \\
ThinkMorph-7B & 46.8 & 46.7 & 45.2 & 47.5 & 48.0 & 49.2 & 48.3 & 48.2 & 43.9 & 46.8 & 48.2 & 49.8 & 46.4 \\
\method (Nav) & 66.5 & 66.5 & 52.9 & 53.2 & 50.2 & 66.8 & 68.1 & 67.4 & 62.3 & 68.2 & 69.5 & 68.2 & 67.3 \\
Prompt-OPSD & 68.2 & 68.2 & 51.8 & 48.3 & 51.3 & 65.1 & 67.4 & 63.7 & -- & 68.8 & 67.8 & 62.0 & 59.5 \\
Noise-OPSD & 58.5 & 58.6 & 52.5 & 47.5 & 48.7 & 60.0 & 61.2 & 55.5 & -- & 58.9 & 58.8 & 57.7 & 54.3 \\
\textbf{\method} & 81.8 & 81.8 & 54.8 & 52.3 & 54.3 & 83.8 & 82.2 & 77.2 & 69.6 & 83.0 & 81.8 & 81.1 & 74.2 \\
\botrule
\end{tabular*}
\end{table*}

\begin{table*}[t]
\caption{Image-condition probe at 512 thinking tokens across mosaic, Gaussian blur, and tile-permutation ladders, including the pretrained BAGEL-7B backbone. scr.: pixels in random order. $\times k$: mosaic with $k$-pixel blocks. $\sigma k$: Gaussian blur with $\sigma=k$ pixels. $g\times g$: permutation of a $g\times g$ grid of intact image tiles. Dashes: not run}\label{tab:probe3}
\scriptsize
\renewcommand{\arraystretch}{0.80}
\setlength{\aboverulesep}{1.2pt}
\setlength{\belowrulesep}{1.2pt}
\setlength{\tabcolsep}{2.5pt}
\begin{tabular*}{\textwidth}{@{\extracolsep\fill}lrrrrrrrrrrrrr@{}}
\toprule
Method & real & scr. & $\times2$ & $\times4$ & $\times8$ & $\times16$ & $\sigma2$ & $\sigma4$ & $\sigma8$ & $2\times2$ & $4\times4$ & $8\times8$ & $16\times16$ \\
\midrule
\multicolumn{14}{@{}l}{\emph{V* (191 items)}} \\
BAGEL-7B & 48.7 & 37.7 & 48.2 & -- & 42.4 & 40.3 & 52.4 & 44.5 & 42.4 & 43.5 & 44.5 & 43.5 & 40.8 \\
ThinkMorph-7B & 34.0 & 36.6 & 35.6 & -- & 34.6 & 36.1 & 36.1 & 35.1 & 35.1 & 37.2 & 36.1 & 37.7 & 36.1 \\
\method (Nav) & 56.0 & 38.2 & 57.6 & -- & -- & 38.2 & 56.5 & 53.9 & -- & 52.9 & 54.5 & 50.8 & 44.0 \\
\textbf{\method} & 67.0 & 34.0 & 66.5 & -- & 61.8 & 46.1 & 66.5 & 56.5 & 46.1 & 57.1 & 57.1 & 53.9 & 47.6 \\
\midrule
\multicolumn{14}{@{}l}{\emph{MMVP (300 items)}} \\
BAGEL-7B & 66.0 & 50.3 & 58.7 & -- & 52.7 & 53.0 & 52.7 & 52.3 & 49.0 & 59.0 & 52.3 & 51.3 & 50.3 \\
ThinkMorph-7B & 44.7 & 45.7 & 45.3 & -- & -- & 47.0 & 51.7 & 46.0 & -- & 48.0 & 47.3 & 49.3 & 51.3 \\
\method (Nav) & 70.0 & 45.3 & 69.3 & -- & 53.0 & 50.3 & 68.7 & 63.3 & 57.7 & 64.7 & 61.0 & 59.3 & 50.0 \\
\textbf{\method} & 78.0 & 46.0 & 74.3 & -- & 53.3 & 52.7 & 72.7 & 63.0 & 62.7 & 75.7 & 63.3 & 59.3 & 53.3 \\
\midrule
\multicolumn{14}{@{}l}{\emph{CV-Bench-2D (1,438 items)}} \\
BAGEL-7B & 62.5 & 36.7 & 58.6 & 57.2 & 53.0 & 47.1 & 59.5 & 56.7 & 53.1 & 46.9 & 42.8 & 42.9 & 42.5 \\
ThinkMorph-7B & 43.0 & 33.5 & 41.7 & 42.6 & 41.9 & 41.2 & 43.1 & 43.0 & 42.4 & 45.0 & 43.4 & 43.3 & 40.6 \\
\method (Nav) & 78.0 & 32.4 & 76.6 & 74.8 & 70.9 & 61.1 & 78.0 & 75.2 & 69.1 & 59.3 & 55.5 & 51.3 & 44.9 \\
\textbf{\method} & 83.4 & 32.5 & 82.8 & 80.7 & 75.2 & 67.8 & 81.2 & 78.4 & 72.7 & 64.0 & 57.4 & 55.1 & 52.4 \\
\midrule
\multicolumn{14}{@{}l}{\emph{CV-Bench-3D (1,200 items)}} \\
BAGEL-7B & 58.8 & 51.4 & -- & 57.1 & 55.6 & 54.3 & 59.5 & 57.8 & 55.0 & 55.8 & 52.8 & 50.0 & 50.9 \\
ThinkMorph-7B & 47.6 & 48.8 & -- & 49.8 & 49.1 & 45.0 & 49.4 & 50.7 & 47.5 & 50.0 & 52.7 & 54.2 & 54.1 \\
\method (Nav) & 71.2 & 50.7 & -- & 73.9 & 72.1 & 65.2 & 72.2 & 70.3 & 69.2 & 68.2 & 63.1 & 57.4 & 55.2 \\
\textbf{\method} & 82.0 & 54.5 & -- & 82.2 & 77.2 & 69.5 & 81.9 & 81.2 & 74.2 & 76.6 & 68.8 & 61.6 & 57.1 \\
\botrule
\end{tabular*}
\end{table*}

\begin{figure*}[t]
\centering
\includegraphics[width=\textwidth]{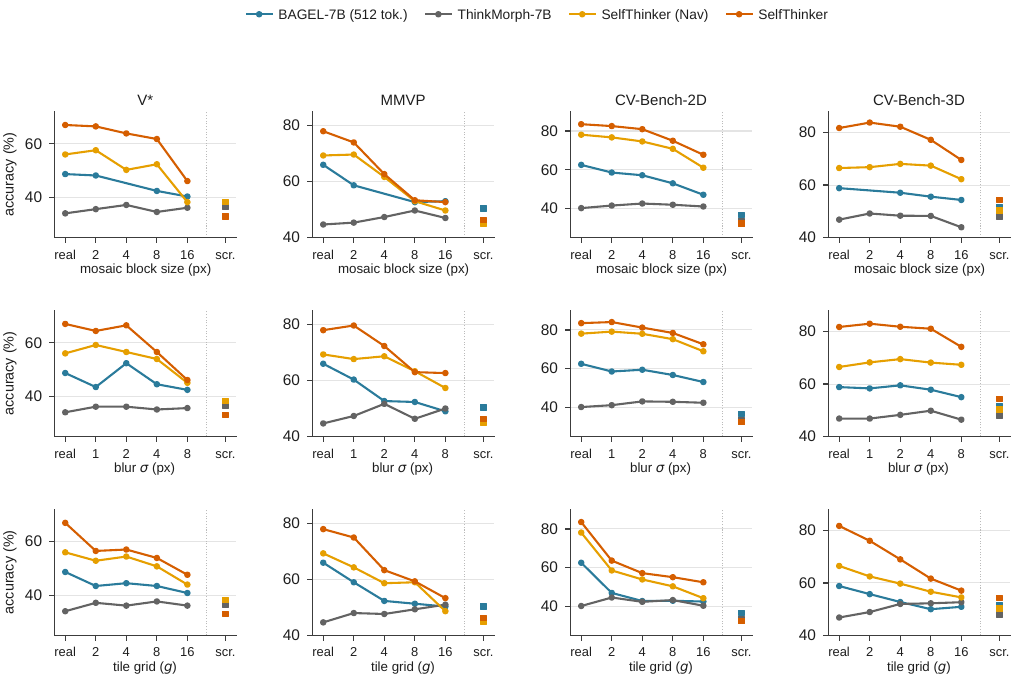}
\caption{Accuracy across mosaic block sizes (top), Gaussian blur $\sigma$ (middle), and $g\times g$ tile permutations (bottom) on V*, MMVP, CV-Bench-2D, and CV-Bench-3D for BAGEL-7B (512 tokens), ThinkMorph-7B, \method (Nav), and \method. Squares on the right mark the pixel-scramble baseline}\label{fig:probe-curves}
\end{figure*}

\section{Context Interventions on Visual Search}\label{app:decode}

We evaluate how ThinkMorph-7B and the \method students respond to five controlled context interventions on the 1,092 verified visual-search items (Table~\ref{tab:intv} and Fig.~\ref{fig:intv}). Whereas ThinkMorph-7B loses 6.1 percentage points when the helper image is removed at test time (45.9\% with ``Own'' versus 39.7\% with ``None''), \emph{\method and label-free \method (Vote) achieve virtually identical accuracy whether given the item's own verified thought (99.0\% and 95.4\%) or no second image at all (98.9\% and 95.3\%, a difference of $+0.1$ points)}. This confirms that the distilled students absorb the privileged visual conditioning without requiring a second image at deployment. Appending a neutral sentence leaves accuracy unchanged (98.9\% and 95.3\%). When the problem image is withheld and only the verified visual thought is shown (``Thought only''), \emph{\method and \method (Vote) achieve 96.5\% and 93.0\% accuracy (compared with 35.4\% for ThinkMorph-7B)}, demonstrating that their understanding pathway directly extracts the solution from the verified thought. Furthermore, to test generalization on unseen problem types, \emph{\method (Nav), trained solely on navigation problems, reaches 92.9\% zero-shot on these 1,092 visual-search items both with and without the visual thought ($0.0$-point gap, up from 45.9\% and 39.7\% for ThinkMorph-7B)}.

\section{Multi-Layer Image-Condition Probe}\label{app:probe}

Tables~\ref{tab:probe1}, \ref{tab:probe2}, and \ref{tab:probe3} report the complete results of the image-condition probe at 256 and 512 thinking tokens. Figures~\ref{fig:probe-examples}, \ref{fig:probe-heat}, and \ref{fig:probe-curves} illustrate the visual transformations, the per-benchmark visual-sensitivity heatmaps, and the three degradation ladders. Unlike the main benchmark evaluation in Table~\ref{tab:main} (which uses 1,024 thinking tokens and semantic grading by Qwen3-30B-A3B-Instruct-2507), the image-condition probe uses a strict 256- or 512-token budget and deterministic rule-based extraction (the last standalone option letter for multiple-choice items or the last number within 5\% relative error for ChartQA) to isolate visual sensitivity under controlled image corruptions. The prompt presents a single image and the question, and multiple-choice items append ``Answer with a single letter.'' Final answers are extracted from explicit \texttt{<answer>} tags or post-\texttt{</think>} text when present, falling back to the last standalone option letter or number before \texttt{<image\_start>}. Image swaps use the previous row of the benchmark file on the subset of items whose previous row depicts a distinct image (222 horizontal and 191 vertical ChartQA items, 1,198 CV-Bench-3D items, and all items elsewhere). Pixel scrambles permute all pixels deterministically, and $g\times g$ tile permutations partition the image into $g^2$ equal rectangular patches and shuffle those patches, preserving local detail inside each patch while disrupting global spatial layout. Mosaics apply box downsampling by factor $k\in\{2,4,8,16\}$ followed by nearest-neighbour upsampling, and blurs apply a Gaussian kernel with $\sigma\in\{1,2,4,8\}$ pixels at native resolution.

\paragraph{Visual perception advantage over teacher controls.}
At 256 thinking tokens (Fig.~\ref{fig:probe-heat} and Tables~\ref{tab:probe1}--\ref{tab:probe2}), \emph{\method achieves the highest accuracy on every benchmark: 67.3\%--68.6\% on ChartQA-h/v, 67.0\% on V*, 78.0\% on MMVP, 83.5\% on CV-Bench-2D, and 81.8\% on CV-Bench-3D}, outperforming ThinkMorph-7B (32.2\%--46.8\%) by wide margins. The real problem image provides a 55--56 point accuracy gain over pixel scramble on ChartQA, 51 points on CV-Bench-2D, 32 points on MMVP, 34 points on V*, and 28 points on CV-Bench-3D (Fig.~\ref{fig:probe-heat}b). Crucially, \emph{\method achieves substantially higher real-image accuracy than both teacher controls (Noise-OPSD and Prompt-OPSD), exceeding them by $+15.7$ and $+16.2$ points on V* (67.0\% versus 51.3\% and 50.8\%), $+4.7$ and $+11.0$ points on MMVP (78.0\% versus 73.3\% and 67.0\%), $+15.2$ and $+10.7$ points on CV-Bench-2D (83.5\% versus 68.3\% and 72.8\%), and $+23.3$ and $+13.6$ points on CV-Bench-3D (81.8\% versus 58.5\% and 68.2\%)}.

\paragraph{Multi-scale spatial perception.}
Comparing the transformation families across Tables~\ref{tab:probe1}--\ref{tab:probe3} and Fig.~\ref{fig:probe-curves} highlights two complementary visual perception capabilities. On ChartQA horizontal and vertical bars (Table~\ref{tab:probe1}), \emph{\method preserves high accuracy on $2\times2$-pixel mosaics (64.9\% and 63.1\%) and $\sigma=1$ blur (68.9\% and 63.4\%)}, demonstrating strong local glyph reading. On V*, MMVP, CV-Bench-2D, and CV-Bench-3D at 512 thinking tokens (Table~\ref{tab:probe3} and Fig.~\ref{fig:probe-curves}), \emph{\method preserves its large advantage over both ThinkMorph-7B and pretrained BAGEL-7B across $2\times2$ to $8\times8$-pixel mosaics, mild blur ($\sigma=2, 4$), and $2\times2$ to $16\times16$ tile permutations}. Specifically, on CV-Bench-2D, \method achieves 82.8\%, 80.7\%, and 75.2\% across $\times2, \times4, \times8$ mosaics (versus 32.5\% on scramble), and on CV-Bench-3D it achieves 82.2\% and 77.2\% across $\times4$ and $\times8$ mosaics (versus 54.5\% on scramble). Together with the \emph{12.0 to 23.2 point improvement over pretrained BAGEL-7B across all four spatial and object-level benchmarks} (Table~\ref{tab:probe3}), these results confirm that \method unlocks high-precision visual perception across both fine local details and global scene geometry.

\end{appendices}

\bibliography{sn-bibliography}

@article{gu2025thinkmorph,
  title   = {{ThinkMorph}: Emergent Properties in Multimodal Interleaved Chain-of-Thought Reasoning},
  author  = {Gu, Jiawei and Hao, Yunzhuo and Wang, Huichen Will and Li, Linjie and Shieh, Michael Qizhe and Choi, Yejin and Krishna, Ranjay and Cheng, Yu},
  journal = {arXiv preprint},
  eprint  = {2510.27492},
  archivePrefix = {arXiv},
  year    = {2025},
  doi     = {10.48550/arXiv.2510.27492}
}

@article{li2026visualopsd,
  title   = {{Visual-OPSD}: Cross-Modal On-Policy Self-Distillation for Efficient Unified Multimodal Reasoning},
  author  = {Li, Pengyu and Gao, Zhitao and Zhang, Lingling and Huang, Muye and Li, Yuanming and Yang, Zesheng and Xu, Fangzhi and Liu, Jun},
  journal = {arXiv preprint},
  eprint  = {2606.18974},
  archivePrefix = {arXiv},
  year    = {2026},
  doi     = {10.48550/arXiv.2606.18974}
}

@article{li2026uopsd,
  title   = {On-Policy Self-Distillation without Any Supervision},
  author  = {Li, Yijiang and Wang, Bingyang and Liang, Yijun and Tian, Yunjie and Fu, Di and Vasconcelos, Nuno},
  journal = {arXiv preprint},
  eprint  = {2608.06296},
  archivePrefix = {arXiv},
  year    = {2026},
  doi     = {10.48550/arXiv.2608.06296}
}

@article{gkountouras2026canon,
  title   = {Consensus as Privileged Context for Label-Free Self-Distillation},
  author  = {Gkountouras, John and Juki{\'c}, Josip and Titov, Ivan},
  journal = {arXiv preprint},
  eprint  = {2607.13643},
  archivePrefix = {arXiv},
  year    = {2026},
  doi     = {10.48550/arXiv.2607.13643}
}

@article{wu2026seed,
  title   = {{SEED}: Self-Evolving On-Policy Distillation for Agentic Reinforcement Learning},
  author  = {Wu, Jinyang and Yang, Shuo and Lu, Zhengxi and Zhang, Fan and Shen, Yuhao and Feng, Lang and Luo, Haoran and Lian, Zheng and Zhang, Shuai and Wen, Zhengqi and others},
  journal = {arXiv preprint},
  eprint  = {2607.14777},
  archivePrefix = {arXiv},
  year    = {2026},
  doi     = {10.48550/arXiv.2607.14777}
}

@article{han2026atesd,
  title   = {Adaptive Teacher Exposure for Self-Distillation in {LLM} Reasoning},
  author  = {Han, Zihao and Zhang, Tiangang and Wang, Huaibin and Sun, Yilun},
  journal = {arXiv preprint},
  eprint  = {2605.11458},
  archivePrefix = {arXiv},
  year    = {2026},
  doi     = {10.48550/arXiv.2605.11458}
}

@article{yuan2026visionopd,
  title   = {{Vision-OPD}: Learning to See Fine Details for Multimodal {LLMs} via On-Policy Self-Distillation},
  author  = {Yuan, Qianhao and Lou, Jie and Yu, Xing and Lin, Hongyu and Sun, Le and Han, Xianpei and Lu, Yaojie},
  journal = {arXiv preprint},
  eprint  = {2605.18740},
  archivePrefix = {arXiv},
  year    = {2026},
  doi     = {10.48550/arXiv.2605.18740}
}

@article{su2025unigame,
  title   = {{UniGame}: Turning a Unified Multimodal Model Into Its Own Adversary},
  author  = {Su, Zhaolong and Lu, Wang and Chen, Hao and Li, Sharon and Wang, Jindong},
  journal = {arXiv preprint},
  eprint  = {2511.19413},
  archivePrefix = {arXiv},
  year    = {2025},
  doi     = {10.48550/arXiv.2511.19413}
}

@article{deng2025bagel,
  title   = {Emerging Properties in Unified Multimodal Pretraining},
  author  = {Deng, Chaorui and Zhu, Deyao and Li, Kunchang and Gou, Chenhui and Li, Feng and Wang, Zeyu and Zhong, Shu and Yu, Weihao and Nie, Xiaonan and Song, Ziang and others},
  journal = {arXiv preprint},
  eprint  = {2505.14683},
  archivePrefix = {arXiv},
  year    = {2025},
  doi     = {10.48550/arXiv.2505.14683}
}

@article{li2025mvot,
  title   = {Imagine while Reasoning in Space: Multimodal Visualization-of-Thought},
  author  = {Li, Chengzu and Wu, Wenshan and Zhang, Huanyu and Xia, Yan and Mao, Shaoguang and Dong, Li and Vuli{\'c}, Ivan and Wei, Furu},
  journal = {arXiv preprint},
  eprint  = {2501.07542},
  archivePrefix = {arXiv},
  year    = {2025},
  doi     = {10.48550/arXiv.2501.07542}
}

@inproceedings{hu2024sketchpad,
  title     = {Visual Sketchpad: Sketching as a Visual Chain of Thought for Multimodal Language Models},
  author    = {Hu, Yushi and Shi, Weijia and Fu, Xingyu and Roth, Dan and Ostendorf, Mari and Zettlemoyer, Luke and Smith, Noah A. and Krishna, Ranjay},
  booktitle = {Advances in Neural Information Processing Systems},
  year      = {2024},
  doi       = {10.48550/arXiv.2406.09403}
}

@article{chern2025thinking,
  title   = {Thinking with Generated Images},
  author  = {Chern, Ethan and Hu, Zhulin and Chern, Steffi and Kou, Siqi and Su, Jiadi and Ma, Yan and Deng, Zhijie and Liu, Pengfei},
  journal = {arXiv preprint},
  eprint  = {2505.22525},
  archivePrefix = {arXiv},
  year    = {2025},
  doi     = {10.48550/arXiv.2505.22525}
}

@article{yang2025machine,
  title   = {Machine Mental Imagery: Empower Multimodal Reasoning with Latent Visual Tokens},
  author  = {Yang, Zeyuan and Yu, Xueyang and Chen, Delin and Shen, Maohao and Gan, Chuang},
  journal = {arXiv preprint},
  eprint  = {2506.17218},
  archivePrefix = {arXiv},
  year    = {2025},
  doi     = {10.48550/arXiv.2506.17218}
}

@article{menon2024whiteboard,
  title   = {Whiteboard-of-Thought: Thinking Step-by-Step Across Modalities},
  author  = {Menon, Sachit and Zemel, Richard and Vondrick, Carl},
  journal = {arXiv preprint},
  eprint  = {2406.14562},
  archivePrefix = {arXiv},
  year    = {2024},
  doi     = {10.48550/arXiv.2406.14562}
}

@article{zheng2025deepeyes,
  title   = {{DeepEyes}: Incentivizing ``Thinking with Images'' via Reinforcement Learning},
  author  = {Zheng, Ziwei and Yang, Michael and Hong, Jack and Zhao, Chenxiao and Xu, Guohai and Yang, Le and Shen, Chao and Yu, Xing},
  journal = {arXiv preprint},
  eprint  = {2505.14362},
  archivePrefix = {arXiv},
  year    = {2025},
  doi     = {10.48550/arXiv.2505.14362}
}

@article{fan2025grit,
  title   = {{GRIT}: Teaching {MLLMs} to Think with Images},
  author  = {Fan, Yue and He, Xuehai and Yang, Diji and Zheng, Kaizhi and Kuo, Ching-Chen and Zheng, Yuting and Narayanaraju, Sravana Jyothi and Guan, Xinze and Wang, Xin Eric},
  journal = {arXiv preprint},
  eprint  = {2505.15879},
  archivePrefix = {arXiv},
  year    = {2025},
  doi     = {10.48550/arXiv.2505.15879}
}

@inproceedings{shao2024visualcot,
  title     = {Visual {CoT}: Advancing Multi-Modal Language Models with a Comprehensive Dataset and Benchmark for Chain-of-Thought Reasoning},
  author    = {Shao, Hao and Qian, Shengju and Xiao, Han and Song, Guanglu and Zong, Zhuofan and Wang, Letian and Liu, Yu and Li, Hongsheng},
  booktitle = {Advances in Neural Information Processing Systems, Datasets and Benchmarks Track},
  year      = {2024},
  doi       = {10.48550/arXiv.2403.16999}
}

@article{chameleon2024,
  title   = {Chameleon: Mixed-Modal Early-Fusion Foundation Models},
  author  = {{Chameleon Team}},
  journal = {arXiv preprint},
  eprint  = {2405.09818},
  archivePrefix = {arXiv},
  year    = {2024},
  doi     = {10.48550/arXiv.2405.09818}
}

@article{zhou2024transfusion,
  title   = {Transfusion: Predict the Next Token and Diffuse Images with One Multi-Modal Model},
  author  = {Zhou, Chunting and Yu, Lili and Babu, Arun and Tirumala, Kushal and Yasunaga, Michihiro and Shamis, Leonid and Kahn, Jacob and Ma, Xuezhe and Zettlemoyer, Luke and Levy, Omer},
  journal = {arXiv preprint},
  eprint  = {2408.11039},
  archivePrefix = {arXiv},
  year    = {2024},
  doi     = {10.48550/arXiv.2408.11039}
}

@inproceedings{xie2025showo,
  title     = {{Show-o}: One Single Transformer to Unify Multimodal Understanding and Generation},
  author    = {Xie, Jinheng and Mao, Weijia and Bai, Zechen and Zhang, David Junhao and Wang, Weihao and Lin, Kevin Qinghong and Gu, Yuchao and Chen, Zhijie and Yang, Zhenheng and Shou, Mike Zheng},
  booktitle = {International Conference on Learning Representations},
  year      = {2025},
  doi       = {10.48550/arXiv.2408.12528}
}

@article{wu2024janus,
  title   = {Janus: Decoupling Visual Encoding for Unified Multimodal Understanding and Generation},
  author  = {Wu, Chengyue and Chen, Xiaokang and Wu, Zhiyu and Ma, Yiyang and Liu, Xingchao and Pan, Zizheng and Liu, Wen and Xie, Zhenda and Yu, Xingkai and Ruan, Chong and Luo, Ping},
  journal = {arXiv preprint},
  eprint  = {2410.13848},
  archivePrefix = {arXiv},
  year    = {2024},
  doi     = {10.48550/arXiv.2410.13848}
}

@article{chen2025januspro,
  title   = {{Janus-Pro}: Unified Multimodal Understanding and Generation with Data and Model Scaling},
  author  = {Chen, Xiaokang and Wu, Zhiyu and Liu, Xingchao and Pan, Zizheng and Liu, Wen and Xie, Zhenda and Yu, Xingkai and Ruan, Chong},
  journal = {arXiv preprint},
  eprint  = {2501.17811},
  archivePrefix = {arXiv},
  year    = {2025},
  doi     = {10.48550/arXiv.2501.17811}
}

@article{wang2024emu3,
  title   = {{Emu3}: Next-Token Prediction is All You Need},
  author  = {Wang, Xinlong and Zhang, Xiaosong and Luo, Zhengxiong and Sun, Quan and Cui, Yufeng and Wang, Jinsheng and Zhang, Fan and Wang, Yueze and Li, Zhen and Yu, Qiying and others},
  journal = {arXiv preprint},
  eprint  = {2409.18869},
  archivePrefix = {arXiv},
  year    = {2024},
  doi     = {10.48550/arXiv.2409.18869}
}

@article{tong2024metamorph,
  title   = {{MetaMorph}: Multimodal Understanding and Generation via Instruction Tuning},
  author  = {Tong, Shengbang and Fan, David and Zhu, Jiachen and Xiong, Yunyang and Chen, Xinlei and Sinha, Koustuv and Rabbat, Michael and LeCun, Yann and Xie, Saining and Liu, Zhuang},
  journal = {arXiv preprint},
  eprint  = {2412.14164},
  archivePrefix = {arXiv},
  year    = {2024},
  doi     = {10.48550/arXiv.2412.14164}
}

@inproceedings{wang2023selfconsistency,
  title     = {Self-Consistency Improves Chain of Thought Reasoning in Language Models},
  author    = {Wang, Xuezhi and Wei, Jason and Schuurmans, Dale and Le, Quoc and Chi, Ed and Narang, Sharan and Chowdhery, Aakanksha and Zhou, Denny},
  booktitle = {International Conference on Learning Representations},
  year      = {2023},
  doi       = {10.48550/arXiv.2203.11171}
}

@inproceedings{zelikman2022star,
  title     = {{STaR}: Bootstrapping Reasoning With Reasoning},
  author    = {Zelikman, Eric and Wu, Yuhuai and Mu, Jesse and Goodman, Noah D.},
  booktitle = {Advances in Neural Information Processing Systems},
  year      = {2022},
  doi       = {10.48550/arXiv.2203.14465}
}

@article{gulcehre2023rest,
  title   = {Reinforced Self-Training ({ReST}) for Language Modeling},
  author  = {Gulcehre, Caglar and Paine, Tom Le and Srinivasan, Srivatsan and Konyushkova, Ksenia and Weerts, Lotte and Sharma, Abhishek and Siddhant, Aditya and Ahern, Alex and Wang, Miaosen and Gu, Chenjie and others},
  journal = {arXiv preprint},
  eprint  = {2308.08998},
  archivePrefix = {arXiv},
  year    = {2023},
  doi     = {10.48550/arXiv.2308.08998}
}

@article{dong2023raft,
  title   = {{RAFT}: Reward rAnked FineTuning for Generative Foundation Model Alignment},
  author  = {Dong, Hanze and Xiong, Wei and Goyal, Deepanshu and Zhang, Yihan and Chow, Winnie and Pan, Rui and Diao, Shizhe and Zhang, Jipeng and Shum, Kashun and Zhang, Tong},
  journal = {Transactions on Machine Learning Research},
  year    = {2023},
  doi     = {10.48550/arXiv.2304.06767}
}

@article{yuan2023scaling,
  title   = {Scaling Relationship on Learning Mathematical Reasoning with Large Language Models},
  author  = {Yuan, Zheng and Yuan, Hongyi and Li, Chengpeng and Dong, Guanting and Lu, Keming and Tan, Chuanqi and Zhou, Chang and Zhou, Jingren},
  journal = {arXiv preprint},
  eprint  = {2308.01825},
  archivePrefix = {arXiv},
  year    = {2023},
  doi     = {10.48550/arXiv.2308.01825}
}

@inproceedings{huang2023selfimprove,
  title     = {Large Language Models Can Self-Improve},
  author    = {Huang, Jiaxin and Gu, Shixiang Shane and Hou, Le and Wu, Yuexin and Wang, Xuezhi and Yu, Hongkun and Han, Jiawei},
  booktitle = {Proceedings of the Conference on Empirical Methods in Natural Language Processing},
  year      = {2023},
  doi       = {10.48550/arXiv.2210.11610}
}

@article{zuo2025ttrl,
  title   = {{TTRL}: Test-Time Reinforcement Learning},
  author  = {Zuo, Yuxin and Zhang, Kaiyan and Sheng, Li and Qu, Shang and Cui, Ganqu and Zhu, Xuekai and Li, Haozhan and Zhang, Yuchen and Long, Xinwei and Hua, Ermo and others},
  journal = {arXiv preprint},
  eprint  = {2504.16084},
  archivePrefix = {arXiv},
  year    = {2025},
  doi     = {10.48550/arXiv.2504.16084}
}

@inproceedings{chen2024spin,
  title     = {Self-Play Fine-Tuning Converts Weak Language Models to Strong Language Models},
  author    = {Chen, Zixiang and Deng, Yihe and Yuan, Huizhuo and Ji, Kaixuan and Gu, Quanquan},
  booktitle = {International Conference on Machine Learning},
  year      = {2024},
  doi       = {10.48550/arXiv.2401.01335}
}

@article{deepseek2025r1,
  title   = {{DeepSeek-R1}: Incentivizing Reasoning Capability in {LLMs} via Reinforcement Learning},
  author  = {{DeepSeek-AI}},
  journal = {arXiv preprint},
  eprint  = {2501.12948},
  archivePrefix = {arXiv},
  year    = {2025},
  doi     = {10.48550/arXiv.2501.12948}
}

@article{hinton2015distilling,
  title   = {Distilling the Knowledge in a Neural Network},
  author  = {Hinton, Geoffrey and Vinyals, Oriol and Dean, Jeff},
  journal = {arXiv preprint},
  eprint  = {1503.02531},
  archivePrefix = {arXiv},
  year    = {2015},
  doi     = {10.48550/arXiv.1503.02531}
}

@article{vapnik2009lupi,
  title   = {A New Learning Paradigm: Learning Using Privileged Information},
  author  = {Vapnik, Vladimir and Vashist, Akshay},
  journal = {Neural Networks},
  volume  = {22},
  number  = {5--6},
  pages   = {544--557},
  year    = {2009},
  doi     = {10.1016/j.neunet.2009.06.042}
}

@inproceedings{lopezpaz2016unifying,
  title     = {Unifying Distillation and Privileged Information},
  author    = {Lopez-Paz, David and Bottou, L{\'e}on and Sch{\"o}lkopf, Bernhard and Vapnik, Vladimir},
  booktitle = {International Conference on Learning Representations},
  year      = {2016},
  doi       = {10.48550/arXiv.1511.03643}
}

@inproceedings{agarwal2024gkd,
  title     = {On-Policy Distillation of Language Models: Learning from Self-Generated Mistakes},
  author    = {Agarwal, Rishabh and Vieillard, Nino and Zhou, Yongchao and Stanczyk, Piotr and Ramos, Sabela and Geist, Matthieu and Bachem, Olivier},
  booktitle = {International Conference on Learning Representations},
  year      = {2024},
  doi       = {10.48550/arXiv.2306.13649}
}

@article{church1990pmi,
  title   = {Word Association Norms, Mutual Information, and Lexicography},
  author  = {Church, Kenneth Ward and Hanks, Patrick},
  journal = {Computational Linguistics},
  volume  = {16},
  number  = {1},
  pages   = {22--29},
  year    = {1990}
}

@inproceedings{koh2017influence,
  title     = {Understanding Black-box Predictions via Influence Functions},
  author    = {Koh, Pang Wei and Liang, Percy},
  booktitle = {International Conference on Machine Learning},
  year      = {2017},
  doi       = {10.48550/arXiv.1703.04730}
}

@article{lanham2023faithfulness,
  title   = {Measuring Faithfulness in Chain-of-Thought Reasoning},
  author  = {Lanham, Tamera and Chen, Anna and Radhakrishnan, Ansh and Steiner, Benoit and Denison, Carson and Hernandez, Danny and Li, Dustin and Durmus, Esin and Hubinger, Evan and Kernion, Jackson and others},
  journal = {arXiv preprint},
  eprint  = {2307.13702},
  archivePrefix = {arXiv},
  year    = {2023},
  doi     = {10.48550/arXiv.2307.13702}
}

@inproceedings{turpin2023unfaithful,
  title     = {Language Models Don't Always Say What They Think: Unfaithful Explanations in Chain-of-Thought Prompting},
  author    = {Turpin, Miles and Michael, Julian and Perez, Ethan and Bowman, Samuel R.},
  booktitle = {Advances in Neural Information Processing Systems},
  year      = {2023},
  doi       = {10.48550/arXiv.2305.04388}
}

@inproceedings{lipman2023flow,
  title     = {Flow Matching for Generative Modeling},
  author    = {Lipman, Yaron and Chen, Ricky T. Q. and Ben-Hamu, Heli and Nickel, Maximilian and Le, Matt},
  booktitle = {International Conference on Learning Representations},
  year      = {2023},
  doi       = {10.48550/arXiv.2210.02747}
}

@inproceedings{liu2023rectified,
  title     = {Flow Straight and Fast: Learning to Generate and Transfer Data with Rectified Flow},
  author    = {Liu, Xingchao and Gong, Chengyue and Liu, Qiang},
  booktitle = {International Conference on Learning Representations},
  year      = {2023},
  doi       = {10.48550/arXiv.2209.03003}
}

@inproceedings{zhai2023siglip,
  title     = {Sigmoid Loss for Language Image Pre-Training},
  author    = {Zhai, Xiaohua and Mustafa, Basil and Kolesnikov, Alexander and Beyer, Lucas},
  booktitle = {Proceedings of the IEEE/CVF International Conference on Computer Vision},
  year      = {2023},
  doi       = {10.48550/arXiv.2303.15343}
}

@article{qwen2025qwen25,
  title   = {{Qwen2.5} Technical Report},
  author  = {{Qwen Team}},
  journal = {arXiv preprint},
  eprint  = {2412.15115},
  archivePrefix = {arXiv},
  year    = {2024},
  doi     = {10.48550/arXiv.2412.15115}
}

@article{yang2025qwen3,
  title   = {{Qwen3} Technical Report},
  author  = {Yang, An and Li, Anfeng and Yang, Baosong and Zhang, Beichen and Hui, Binyuan and Zheng, Bo and Yu, Bowen and Gao, Chang and Huang, Chengen and Lv, Chenxu and others},
  journal = {arXiv preprint},
  eprint  = {2505.09388},
  archivePrefix = {arXiv},
  year    = {2025},
  doi     = {10.48550/arXiv.2505.09388}
}

@inproceedings{kwon2023vllm,
  title     = {Efficient Memory Management for Large Language Model Serving with {PagedAttention}},
  author    = {Kwon, Woosuk and Li, Zhuohan and Zhuang, Siyuan and Sheng, Ying and Zheng, Lianmin and Yu, Cody Hao and Gonzalez, Joseph E. and Zhang, Hao and Stoica, Ion},
  booktitle = {Proceedings of the ACM Symposium on Operating Systems Principles},
  year      = {2023},
  doi       = {10.48550/arXiv.2309.06180}
}

@inproceedings{duan2024vlmevalkit,
  title     = {{VLMEvalKit}: An Open-Source Toolkit for Evaluating Large Multi-Modality Models},
  author    = {Duan, Haodong and Fang, Xinyu and Yang, Junming and Zhao, Xiangyu and Ma, Zerun and Qiao, Yuxuan and Li, Mo and Liang, Tianhao and Zhu, Lin and Agarwal, Amit and others},
  booktitle = {Proceedings of the ACM International Conference on Multimedia},
  year      = {2024},
  doi       = {10.48550/arXiv.2407.11691}
}

@article{wu2024vsp,
  title   = {{VSP}: Assessing the Dual Challenges of Perception and Reasoning in Spatial Planning Tasks for {VLMs}},
  author  = {Wu, Qiucheng and Zhao, Handong and Saxon, Michael and Bui, Trung and Wang, William Yang and Zhang, Yang and Chang, Shiyu},
  journal = {arXiv preprint},
  eprint  = {2407.01863},
  archivePrefix = {arXiv},
  year    = {2024},
  doi     = {10.48550/arXiv.2407.01863}
}

@inproceedings{masry2022chartqa,
  title     = {{ChartQA}: A Benchmark for Question Answering about Charts with Visual and Logical Reasoning},
  author    = {Masry, Ahmed and Long, Do Xuan and Tan, Jia Qing and Joty, Shafiq and Hoque, Enamul},
  booktitle = {Findings of the Association for Computational Linguistics},
  year      = {2022},
  doi       = {10.48550/arXiv.2203.10244}
}

@inproceedings{wu2024vstar,
  title     = {{V*}: Guided Visual Search as a Core Mechanism in Multimodal {LLMs}},
  author    = {Wu, Penghao and Xie, Saining},
  booktitle = {Proceedings of the IEEE/CVF Conference on Computer Vision and Pattern Recognition},
  year      = {2024},
  doi       = {10.48550/arXiv.2312.14135}
}

@inproceedings{fu2024blink,
  title     = {{BLINK}: Multimodal Large Language Models Can See but Not Perceive},
  author    = {Fu, Xingyu and Hu, Yushi and Li, Bangzheng and Feng, Yu and Wang, Haoyu and Lin, Xudong and Roth, Dan and Smith, Noah A. and Ma, Wei-Chiu and Krishna, Ranjay},
  booktitle = {Proceedings of the European Conference on Computer Vision},
  year      = {2024},
  doi       = {10.48550/arXiv.2404.12390}
}

@inproceedings{tong2024eyes,
  title     = {Eyes Wide Shut? Exploring the Visual Shortcomings of Multimodal {LLMs}},
  author    = {Tong, Shengbang and Liu, Zhuang and Zhai, Yuexiang and Ma, Yi and LeCun, Yann and Xie, Saining},
  booktitle = {Proceedings of the IEEE/CVF Conference on Computer Vision and Pattern Recognition},
  year      = {2024},
  doi       = {10.48550/arXiv.2401.06209}
}

@article{ray2024sat,
  title   = {{SAT}: Dynamic Spatial Aptitude Training for Multimodal Language Models},
  author  = {Ray, Arijit and Duan, Jiafei and Brown, Ellis and Tan, Reuben and Bashkirova, Dina and Hendrix, Rose and Ehsani, Kiana and Kembhavi, Aniruddha and Plummer, Bryan A. and Krishna, Ranjay and others},
  journal = {arXiv preprint},
  eprint  = {2412.07755},
  archivePrefix = {arXiv},
  year    = {2024},
  doi     = {10.48550/arXiv.2412.07755}
}

@inproceedings{tong2024cambrian,
  title     = {{Cambrian-1}: A Fully Open, Vision-Centric Exploration of Multimodal {LLMs}},
  author    = {Tong, Shengbang and Brown, Ellis and Wu, Penghao and Woo, Sanghyun and Middepogu, Manoj and Akula, Sai Charitha and Yang, Jihan and Yang, Shusheng and Iyer, Adithya and Pan, Xichen and others},
  booktitle = {Advances in Neural Information Processing Systems},
  year      = {2024},
  doi       = {10.48550/arXiv.2406.16860}
}

@article{towers2024gymnasium,
  title   = {Gymnasium: A Standard Interface for Reinforcement Learning Environments},
  author  = {Towers, Mark and Kwiatkowski, Ariel and Terry, Jordan and Balis, John U. and De Cola, Gianluca and Deleu, Tristan and Gou{\~a}o, Manuel and Kallinteris, Andreas and Krimmel, Markus and KG, Arjun and others},
  journal = {arXiv preprint},
  eprint  = {2407.17032},
  archivePrefix = {arXiv},
  year    = {2024},
  doi     = {10.48550/arXiv.2407.17032}
}

@article{shepard1971mental,
  title   = {Mental Rotation of Three-Dimensional Objects},
  author  = {Shepard, Roger N. and Metzler, Jacqueline},
  journal = {Science},
  volume  = {171},
  number  = {3972},
  pages   = {701--703},
  year    = {1971},
  doi     = {10.1126/science.171.3972.701}
}

@article{kosslyn2001neural,
  title   = {Neural Foundations of Imagery},
  author  = {Kosslyn, Stephen M. and Ganis, Giorgio and Thompson, William L.},
  journal = {Nature Reviews Neuroscience},
  volume  = {2},
  number  = {9},
  pages   = {635--642},
  year    = {2001},
  doi     = {10.1038/35090055}
}

@article{pearson2015mental,
  title   = {Mental Imagery: Functional Mechanisms and Clinical Applications},
  author  = {Pearson, Joel and Naselaris, Thomas and Holmes, Emily A. and Kosslyn, Stephen M.},
  journal = {Trends in Cognitive Sciences},
  volume  = {19},
  number  = {10},
  pages   = {590--602},
  year    = {2015},
  doi     = {10.1016/j.tics.2015.08.003}
}

@article{hanley1982auc,
  title   = {The Meaning and Use of the Area under a Receiver Operating Characteristic ({ROC}) Curve},
  author  = {Hanley, James A. and McNeil, Barbara J.},
  journal = {Radiology},
  volume  = {143},
  number  = {1},
  pages   = {29--36},
  year    = {1982},
  doi     = {10.1148/radiology.143.1.7063747}
}

@inproceedings{chen2024mmstar,
  title     = {Are We on the Right Way for Evaluating Large Vision-Language Models?},
  author    = {Chen, Lin and Li, Jinsong and Dong, Xiaoyi and Zhang, Pan and Zang, Yuhang and Chen, Zehui and Duan, Haodong and Wang, Jiaqi and Qiao, Yu and Lin, Dahua and Zhao, Feng},
  booktitle = {Advances in Neural Information Processing Systems},
  year      = {2024},
  doi       = {10.48550/arXiv.2403.20330}
}

@inproceedings{rahmanzadehgervi2024blind,
  title     = {Vision Language Models Are Blind},
  author    = {Rahmanzadehgervi, Pooyan and Bolton, Logan and Taesiri, Mohammad Reza and Nguyen, Anh Totti},
  booktitle = {Proceedings of the Asian Conference on Computer Vision},
  pages     = {18--34},
  year      = {2024},
  doi       = {10.48550/arXiv.2407.06581}
}

@inproceedings{hendrycks2019robustness,
  title     = {Benchmarking Neural Network Robustness to Common Corruptions and Perturbations},
  author    = {Hendrycks, Dan and Dietterich, Thomas},
  booktitle = {International Conference on Learning Representations},
  year      = {2019},
  doi       = {10.48550/arXiv.1903.12261}
}

@article{askell2021general,
  title   = {A General Language Assistant as a Laboratory for Alignment},
  author  = {Askell, Amanda and Bai, Yuntao and Chen, Anna and Drain, Dawn and Ganguli, Deep and Henighan, Tom and Jones, Andy and Joseph, Nicholas and Mann, Ben and DasSarma, Nova and others},
  journal = {arXiv preprint},
  eprint  = {2112.00861},
  archivePrefix = {arXiv},
  year    = {2021},
  doi     = {10.48550/arXiv.2112.00861}
}

@article{snell2022context,
  title   = {Learning by Distilling Context},
  author  = {Snell, Charlie and Klein, Dan and Zhong, Ruiqi},
  journal = {arXiv preprint},
  eprint  = {2209.15189},
  archivePrefix = {arXiv},
  year    = {2022},
  doi     = {10.48550/arXiv.2209.15189}
}

@article{chen2025sifthinker,
  title={SIFThinker: Spatially-Aware Image Focus for Visual Reasoning},
  author={Chen, Zhangquan and Zhao, Ruihui and Luo, Chuwei and Sun, Mingze and Yu, Xinlei and Kang, Yangyang and Huang, Ruqi},
  journal={arXiv preprint arXiv:2508.06259},
  year={2025}
}

@article{chen2025visrl,
  title={Visrl: Intention-driven visual perception via reinforced reasoning},
  author={Chen, Zhangquan and Luo, Xufang and Li, Dongsheng},
  journal={arXiv preprint arXiv:2503.07523},
  year={2025}
}

@article{chen20264dthinker,
  title={4DThinker: Thinking with 4D Imagery for Dynamic Spatial Understanding},
  author={Chen, Zhangquan and Zhang, Manyuan and Yu, Xinlei and An, Xiang and Li, Bo and Xie, Xin and Wang, ZiDong and Sun, Mingze and Chen, Shuang and Li, Hongyu and others},
  journal={arXiv preprint arXiv:2605.05997},
  year={2026}
}

@article{chen20253dthink,
  title={Think with 3D: Geometric Imagination Grounded Spatial Reasoning from Limited Views},
  author={Chen, Zhangquan and Zhang, Manyuan and Yu, Xinlei and Luo, Xufang and Sun, Mingze and Pan, Zihao and Feng, Yan and Pei, Peng and Cai, Xunliang and Huang, Ruqi},
  journal={arXiv preprint arXiv:2510.18632},
  year={2025}
}

\end{document}